\documentclass[acmlarge]{acmart}
\graphicspath{ {figures/} }
\usepackage{multirow}
\usepackage{subcaption}
\usepackage{threeparttable}
\usepackage{xcolor}

\newcommand{\rom}[1]{\lowercase\expandafter{\romannumeral #1\relax}}

\AtBeginDocument{%
  \providecommand\BibTeX{{%
    \normalfont B\kern-0.5em{\scshape i\kern-0.25em b}\kern-0.8em\TeX}}}

\setcopyright{acmcopyright}
\copyrightyear{2020}
\acmYear{2020}
\acmDOI{10.1145/1122445.1122456}

\acmJournal{IMWUT}
\acmVolume{1}
\acmNumber{2}
\acmArticle{57}
\acmMonth{6}

\usepackage{mycomments3}
\begin{document}

\title{Weakly Supervised Multi-Task Representation Learning for Human Activity Analysis Using Wearables}


\author{Taoran Sheng}
\affiliation{%
  \institution{The University of Texas at Arlington}
  \streetaddress{500 UTA Blvd}
  \city{Arlington}
  \state{Texas}
  \country{USA}
  \postcode{76019-0015}}
\email{taoran.sheng@mavs.uta.edu}

\author{Manfred Huber}
\affiliation{%
  \institution{The University of Texas at Arlington}
  \streetaddress{500 UTA Blvd}
  \city{Arlington}
  \state{Texas}
  \country{USA}
  \postcode{76019-0015}}
\email{huber@cse.uta.edu}

%\affiliation{%
%  \institution{Inria Paris-Rocquencourt}
%\affiliation{%
% \institution{Rajiv Gandhi University}
%\author{H<uifen Chan}
%\affiliation{%
%  \institution{Tsinghua University}
%  \streetaddress{30 Shuangqing Rd}
%\affiliation{%
%  \institution{Palmer Research Laboratories}
%  \streetaddress{8600 Datapoint Drive}
%\affiliation{\institution{The Th{\o}rv{\"a}ld Group}}
%\email{jsmith@affiliation.org}
%\affiliation{\institution{The Kumquat Consortium}}
%\email{jpkumquat@consortium.net}

\renewcommand{\shortauthors}{Sheng et al.}

\begin{abstract}
Sensor data stream\add{s} from wearable devices and smart environments are widely studied in areas like human activity recognition (HAR), person identification, \mdel{etc}\madd{or health monitoring}. However, most of the previous works \madd{in activity and sensor stream analysis} have been focusing on one aspect of the data, \mdel{i.e.}\madd{e.g.} only recognizing the type of the activity or only identifying the person who performed the activity. We instead propose an approach that uses a weakly supervised multi-output siamese network that learns to map the data into multiple representation spaces, where each representation space focuses on one aspect of the data. The representation vectors of the data samples are positioned in the space such that the data with the same semantic meaning in that aspect are closely located to each other. Therefore, as demonstrated with \del{the}\add{a set of} experiments, the trained model can provide metrics for clustering data based on multiple aspects, allowing it to address multiple tasks simultaneously and even to outperform single task supervised methods in many situations. \hil{In addition, \mmdel{a series of} further experiments\mmdel{,} \mmadd{are presented} that \mmadd{in more detail} analyze the effect of \mmadd{the architecture and of using multiple tasks} within this \mmdel{the} \mmdel{multi-task} framework, \mmadd{that investigate the scalability of the model to include additional tasks, and that demonstrate the ability of the framework to combine data for which only partial relationship information with respect to the target tasks is available.}\mmdel{show the capability of the model learning with partially observable data, and assess the scalability of the model to include other new learning task, are conducted for a comprehensive evaluation of the proposed method.}}

\com{Do you want to keep this very broad - i.e. outside activity recognition - or narrow it a little in the beginning ?}
\end{abstract}



\setcopyright{acmcopyright}
\acmJournal{IMWUT}
\acmYear{2020} \acmVolume{4} \acmNumber{2} \acmArticle{57} \acmMonth{6} \acmPrice{15.00}\acmDOI{10.1145/3397330}


\maketitle

\section{Introduction}
The increased availability of sensors embedded in the environment and worn on the body opens up a vast potential to improve many aspects of life and the workplace, including health management, assistive technology, and workplace safety and efficiency. To realize this potential, it is essential to address the arising need for improved human activity analysis from the sensor data stream. Many different methods have been developed to address this need. While current machine learning methods have achieved impressive results, \mdel{however,} most of these methods have limitations in the following perspectives: (\rom{1}) The methods focus on one aspect of the data by either only recognizing the activity \cite{cnnHAR} or only identifying the person who performed the activity \cite{cbbi}; (\rom{2}) Most current person identification methods are based on iris, face, fingerprint or gait identification, in which specific input or activity is needed from the user side \cite{faceVoice,Gafurov2009}; (\rom{3}) Pure supervised training of the model is used which \mdel{needs}\madd{requires} large amounts of labeled data \cite{cnnLSTMHAR}, \mdel{which}\madd{that} is often hard to come by, especially in personalized applications; \hil{(\rom{4}) Most \mmdel{of the} previous methods rely heavily on \mmadd{data that is} \mmdel{completely} \mmadd{labeled with respect to all aspects of the task and thus make it difficult to combine datasets that were created for different purposes, even if they were collected in the same environment, since each dataset would only contain the labels for that specific task. As a result\mmmadd{,} significant amounts of costly re-labeling would be required which might not even be possible in many real world scenarios since the required information for the labeling is no longer available in the data.} \mmdel{observable data, unfortunately, in many real world scenarios, the data is only partially visible to the model.}}

This paper presents an approach that attempts to mitigate these limitations. Intuitively, different persons perform activities in different ways. Distinct personal characteristics are commonly present in all the activities performed by the person. Thus, identifying a person with different types of activities can generalize the model to activity-based identification which does\mdel{n't} \madd{not} need a specific user\mmadd{'s} cooperation. Moreover, from the multi-task\mmdel{ing} perspective, sharing knowledge between related tasks leads to learning generalized representations, helping to reduce the risk of overfitting one specific task. Since \madd{human activity recognition (}HAR\madd{)} and person identification are closely related, combining them into one multi-task model is reasonable and can be beneficial for both tasks. Furthermore, while learning of a selective representation is commonly achieved by supervised training, previous works in computer vision \cite{Koch2015SiameseNN,faceverification} have shown the possibility to achieve similar performance in a weakly supervised manner while dramatically reducing the effort needed to obtain training data.\com{Maybe there should be a reference here ?}

\mdel{Therefore}\madd{Based on these observations},\com{I am not sure if I like this. Maybe we should just leave it out.} this paper proposes a unified deep learning architecture centered around siamese networks and temporal convolutions for simultaneous HAR and activity-based person identification, which is trained \mdel{only} using \madd{only} the information about the similarity of the activities and the persons without knowing the explicit labels. \hil{\mmadd{Moreover, the developed architecture is non task-specific, easily extendable to larger numbers of tasks, and can be trained with data where only partial similarity information is available, i.e. where each data item only contains relationship information for a subset of the tasks that the system is to be trained for.} 

\mmadd{To demonstrate the potential and scalability of the proposed model and to evaluate its effectiveness in different scenarios, a number of experiments are presented and analyzed: (\rom{1}) To evaluate \mmmadd{the proposed method}, the framework is applied to the HAR and person identification problem and compared to existing single and multi-task approaches in these domains. The} \mmdel{ experiment} results and visualization \madd{of the formed representation space} confirm that our method can learn \mmdel{both tasks}\mmadd{them} successfully\mmdel{.}\mmadd{, and comparison with state-of-the-art methods\madd{, including fully supervised methods that take advantage of additional, explicit labels,} demonstrate that our method can achieve competitive performance} on several datasets; \mmdel{

Additionally, we further perform a set of experiments to evaluate the effectiveness of the proposed method in different scenarios: }(\mmdel{\rom{1}}\mmadd{\rom{2}}) \mmdel{In order t}\mmadd{T}o assess the scalability of the proposed method, we expand the model to learn representations for additional attributes in the data along with HAR and person identification\mmadd{. These experiments show that the proposed architecture can scale to increasing numbers of tasks with very little loss of accuracy;} (\mmdel{\rom{2}}\mmadd{\rom{3}}) Ablation studies are provided to investigate the influence of multi-task learning \mmadd{within the architecture, demonstrating not only that multiple tasks can be learned but that training for multiple tasks often increases the performance for individual tasks}; (\mmdel{\rom{3}}\mmadd{\rom{4}}) \mmadd{To investigate the impact of data that has only partial similarity information, an experiment is performed where each data item only contains similarity information with respect to one of the tasks and only to a subset of the data. The results here show that the proposed model can successfully learn all tasks even with partial similarity information with a performance that is close to the one with full information.} \mmdel{The proposed method is tested by training with partially observable data, which is a more difficult and realistic situation than training with completely observable data. Those experiments show that our method is not restricted to any specific task, can be easily expanded to other related task\add{s}, and is capable of learning with incompletely observable data.} \mmdel{Moreover, the experiments also demonstrate that our method can achieve competitive performance compared to state-of-the-art methods \madd{, including fully supervised methods that take advantage of additional, explicit labels,} on several datasets.}}

\section{Related Work}
\subsection{HAR and Person Identification}
%HAR and person identification include vision-based, and sensor-based methods. We focus on the sensor-based methods, more specifically, wearable sensor-based methods. Both HAR and person identification can be considered as a classification problem, where the input is the sensor data stream and the output is the corresponding activity label and person label. 

%\subsubsection{HAR}
There are two main directions in wearable sensor-based HAR, either handcrafted feature\mmmadd{-}based methods or deep neural network (DNN) based methods. Handcrafted features are designed with domain knowledge. For example, \cite{sbhar,transitionAware} used statistical features, \mmdel{e.g.}\mmadd{such as}\mdel{,} mean, variance and entropy, in their models. Features extracted from \madd{a} wavelet transform were utilized in \cite{waveletFeature}. He and Jin \cite{dctFeature} used features extracted by applying discrete cosine transform. The advantage of these features is that they can be derived from the signal easily and have been shown to be effective in the HAR system. However, domain knowledge is required to design the features manually.
Recently, many HAR models adopt DNN to allow automatic feature extraction. Morales and Roggen \cite{cnnLSTMHAR} proposed a model consisting of convolutional neural network (CNN) and long short-term memory (LSTM) components. CNN is used here to capture local temporal relations while the memory states of LSTM ease the learning of long time scale dependencies. \add{I}n \cite{DBNHAR}, a hybrid approach was proposed, which used a deep belief network as an emission matrix of a \mmdel{h}\mmadd{H}idden Markov \mmdel{m}\mmadd{M}odel to model the sequence of human activities. These methods can automatically extract features from the data without any domain knowledge. But, they still require explicit labels to supervise the training of the model.

%\subsubsection{Person Identification}
Many person identification methods are based on iris, face, and fingerprint. Those kinds of methods require specific cooperation or explicit action/input from the user side, e.g.\mdel{,} standing in front of a camera, looking at a specific point\madd{,} etc\madd{.} \cite{piDailyAct}, while gait recognition-based person identification, e.g.\mdel{,} \cite{Gafurov2009,1415569,gafhi}, enables an inexpensive, convenient, and unobtrusive way to complete the task. However, gait-based methods assume that walking is the only activity to be performed during the identification. In many real-world applications, this assumption may not hold.
There have been only a few studies that are identifying the person based on various activities recorded by sensors. Kwapisz, Weiss, and Moore \cite{cbbi} proposed a model which used handcrafted features and decision trees. Their model addressed the biometric identification task by analyzing four types of dynamic activities (walking, jogging, ascending and descending stairs). Elkader et al \cite{piDailyAct} expanded the person identification method from \madd{a} limited number of specific activities to a set of various normal daily activities. However, these works still focus on only one aspect of the data.

The most \mdel{relevant}\madd{closely related} works are perhaps \cite{piPositionKinect,bioinsight}. In \cite{piPositionKinect}, Reddy et al. proposed a method to first identify the person's states: standing, sitting, or walking, then separate SVM based models are used for identifying the person for each of the three mentioned states. \hil{This method addresses posture recognition and person identification, but the two tasks are addressed in two separate steps by separate models. Hernandez, McDuff, and Picard \cite{bioinsight} collected \mmdel{the} experimental data from a wrist worn smartwatch, and their proposed model identifies the person and three static body postures (sitting, standing, and lying) at the same time. These two works address two tasks, but \mmadd{while the first uses walking for activity recognition, in both cases} the\mmdel{ir} methods \mmadd{for person identification} are based \mmadd{solely} on a very small number of simple static postures, and no \mmadd{information about the} dynamic\mmadd{s of}  activities \mmdel{are}\mmadd{is used for this task}\mmdel{ included in the \mmdel{activity recognition and} person identification tasks}.  Hence, these two methods \mmadd{do not support}\mmdel{are not} general activity-based \mmadd{person} identification\mmdel{ as well}. \mmadd{Moreover, both use a very small set of activities and are not easily expandable to a more general set, and do not provide any explicit means to add additional attributes or tasks.} \mmadd{In contrast, the framework presented here does not make any assumptions about the type of activities and provides a uniform, consistent model for multi-task learning of both static and dynamic attributes. For this, it develops a novel siamese network training architecture that allows efficient training of a latent representation that accurately embeds the chosen attributes. In addition, it extends the state-of-the art by permitting the system to be trained without explicit labels, just using similarity or relationship information, as well as to be trained with datasets that only contain partial similarity information.}}

\subsection{Siamese Networks and Temporal Convolution Networks}
We take inspiration from siamese architectures\mmdel{,} and temporal convolutional networks (TCN) to design our model that can efficiently capture the temporal patterns and compute the semantic similarity between the pairs of data sequences. 

\begin{figure}
\begin{center}
\includegraphics[width=0.55\textwidth]{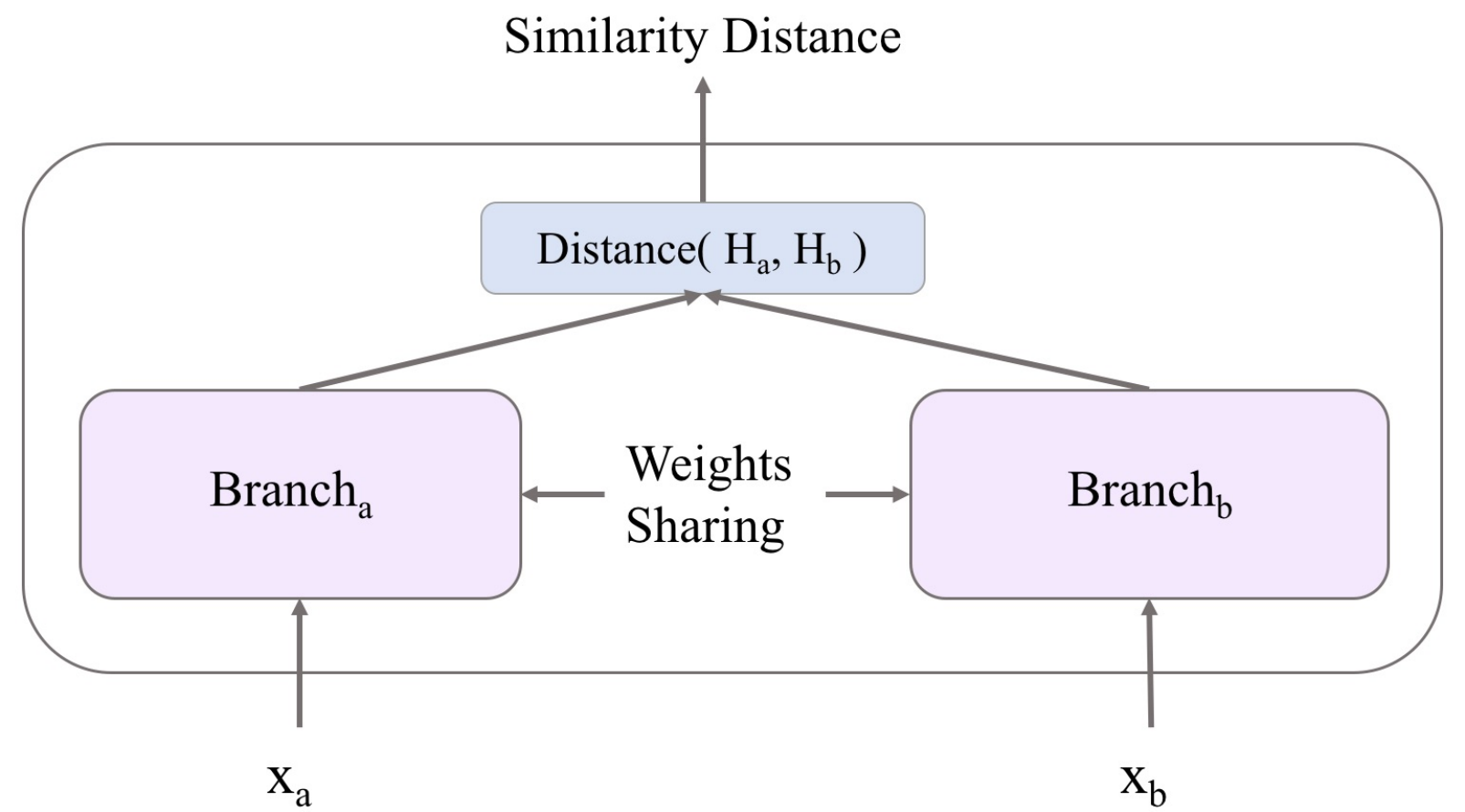}
\caption{The basic architecture of a siamese network.} \label{siamgnrl}
\end{center}
\end{figure}

%\subsubsection{Siamese Networks}
\mdel{The}\madd{A} siamese network \cite{signature1993} is a neural network with dual branches and shared weights. As illustrated in Fig.~\ref{siamgnrl}, it processes an input pair $\{x_a, x_b\}$ and yields a pair of comparable representation vectors $\{H_a, H_b\}$. The distance between the comparable representation vectors is then used as the semantic similarity of the input pair. The siamese architecture is widely used in many domains. Originally \cite{signature1993}, a siamese network \mdel{is}\madd{was} used to verify signatures. Mueller and Thyagarajan \cite{sentenceSiamese} used a siamese recurrent neural network\mmmdel{s} to measure the semantic similarity between a pair of sentences. In \cite{faceverification}, a siamese CNN is proposed to learn a complex similarity metric for face verification. In other areas, the siamese architecture has been applied in unsupervised acoustic model learning \cite{jointlearning,Synnaeve2016ATC,Kamper2015DeepCA}, image recognition \cite{eccvsiamese} and object tracking \cite{Koch2015SiameseNN}.

\begin{figure}
\begin{center}
\includegraphics[width=0.55\textwidth]{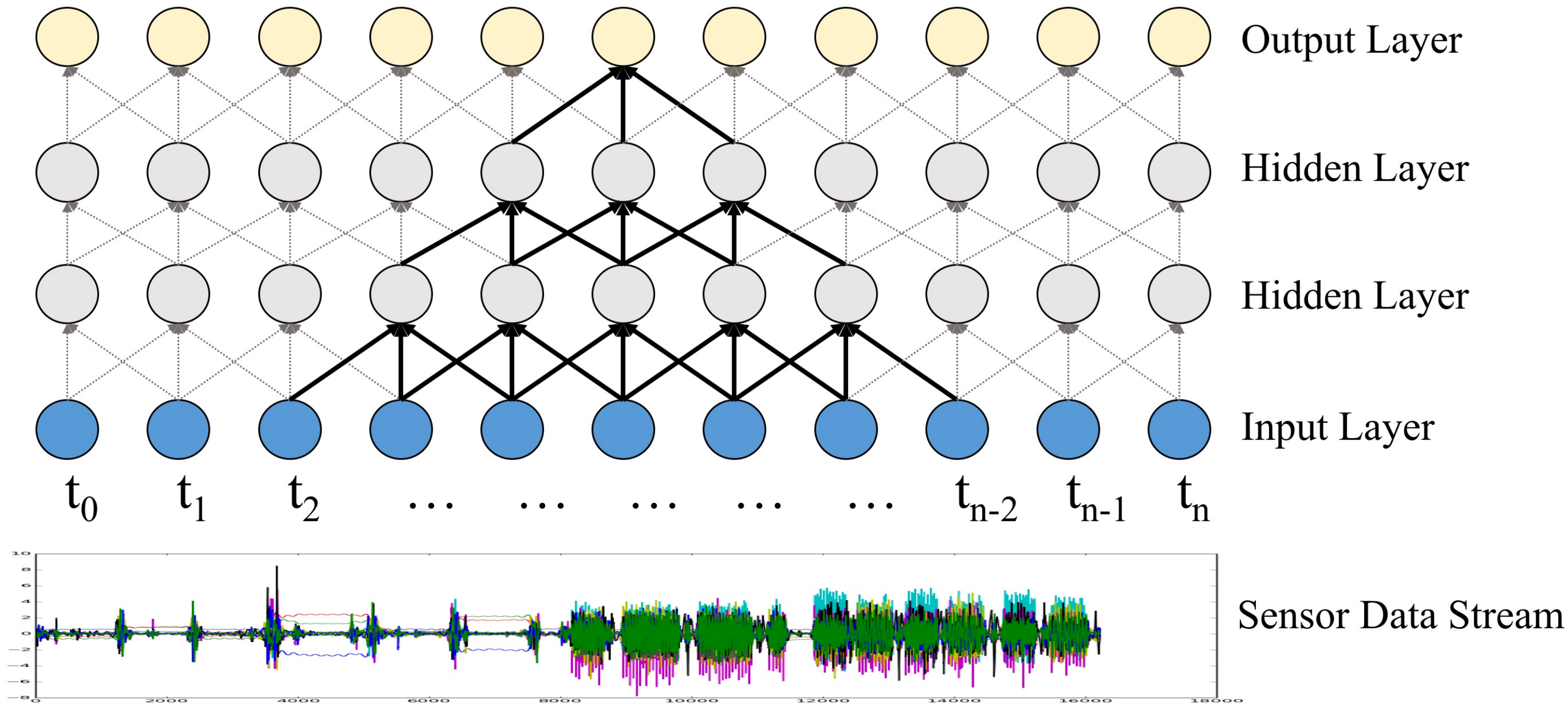}
\caption{The basic architecture of a temporal convolutional network \madd{(TCN)}.} \label{TCNgnrl}
\end{center}
\end{figure}

%\subsubsection{Temporal Convolutional Networks}
\hil{A TCN uses a hierarchy of convolutions to abstract the temporal relations of the data stream at different time scales. As shown in Fig.~\ref{TCNgnrl}, the hierarchical structure of the TCN is effective to learn the \mmadd{incrementally} long\mmadd{er}-range temporal patterns in the data stream \mmadd{with each increase in the depth of the network}. More specifically, the low-level TCN layers focus on capturing simple features within a short period of time, while the high-level layers learn to aggregate the low-level simple features into more complex and abstract high-level concepts within a long\mmadd{er} period of time. Hence, stacking TCN layers together is an efficient way to model the long\mmadd{er}-range temporal patterns in the data.} TCN has been successfully applied in many different areas, e.g., computer vision (video data stream), natural language processing (speech or text data)\madd{,} etc. In \cite{wavenet}, Oord et al. proposed a model that can generate raw speech signals. In \cite{TCN1,TCN2}, TCN has been used to segment and detect action in the video. In \cite{charTCN}, a character-level TCN is used to classify text.

\section{Proposed Method}
Our work differs from previous works in several important ways. First, unlike most of the previous methods, our proposed model can identify the activity and the person simultaneously and can be systematically expanded to identify additional types of attributes. Second, in comparison to most of the previous wearable sensor-based person identification methods, which have been customized to either only focused on gait information, or only used a limited number of dynamic activities or static body postures, our model uses a task-agnostic structure and is \madd{here} evaluated on four diverse public datasets \cite{pamap2,mhealth,transitionAware,wisdm}. These datasets, respectively, contain 12, 12, 7, or 6 types of different activities performed by 9, 10, 30, or 36 different persons. This evaluation offers a more realistic and challenging scenario, and through the use of a wider range of activities may lead to a more robust identification model. Moreover, since in the daily routine of a person in real life many different activities will be performed, an identification system that is restricted to a very small number of activities might fail to work, and even in situations where it can still make correct classifications can only take advantage of a small part of the available data. \hil{Third, our proposed method is capable of learning with \mmdel{partially observable} data \mmadd{with only partial similarity or data relationship information}, which could be useful in many real-world applications \mmadd{where}\mmdel{that a perfect observation of} \mmadd{not all attributes of the data can be determined for any given point in time or where multiple datasets that were collected for different purposes using the same sensors in the same domain\mmmdel{ but for different purposes} can be combined to obtain a larger training set and consequently better task performance and less risk of overfitting}\mmdel{the data is not available}. Finally, our method, by taking advantage of the siamese architecture, can be trained in a weakly supervised manner, hence no explicit labels are needed.}

To achieve these capabilities, we frame the problem as learning an invariant mapping that maps the input data sequence into a semantic representation space. The learning process relies only on the relationship of the data sequences in the input pair, therefore the learned model will map two data sequences either to the same area in the representation space if the two data sequences are semantically similar, or to different areas if the data sequences are semantically dissimilar.
Formally, given a pair of sensor data sequences $\{x_a, x_b\}$, the aim is to learn a mapping $f$ that maps the pair $\{x_a, x_b\}$ into a representation space such that:
\begin{align}
& H_{x_a} = f(x_a) \\
& H_{x_b} = f(x_b)
\end{align}

The distance between the representation pair $\{H_{x_a}, H_{x_b}\}$ approximates the semantic similarity of the input pair $\{x_a, x_b\}$. Specifically, our proposed model learns two such mappings: 
\begin{equation}
  f^{act}:\begin{cases}
    x_a \rightarrow H^{act}_{x_a} \\
    x_b \rightarrow H^{act}_{x_b}
  \end{cases}
\end{equation}

\begin{equation}
  f^{pers}:\begin{cases}
    x_a \rightarrow H^{pers}_{x_a} \\
    x_b \rightarrow H^{pers}_{x_b}
  \end{cases}
\end{equation}

The mapping $f^{act}$ is based on the activity similarity of $x_a$ and $x_b$, such that if $x_a$ and $x_b$ belong to the same type of activity, then $H^{act}_{x_a}$ and $H^{act}_{x_b}$ will stay together, while $H^{act}_{x_a}$ and $H^{act}_{x_b}$ will stay away from each other, if $x_a$ and $x_b$ are from different types of activities. The mapping $f^{pers}$ is based on the performer of the activity, such that if $x_a$ and $x_b$ are performed by the same person, then $H^{pers}_{x_a}$ and $H^{pers}_{x_b}$ will stay together, while $H^{pers}_{x_a}$ and $H^{pers}_{x_b}$ will stay away from each other\mdel{,} if $x_a$ and $x_b$ are performed by different persons. The mappings $f^{act}$ and $f^{pers}$ preserve different semantic relationships between the input data sequences. The proposed model learns these two mappings at the same time.

\subsection{TCN Blocks}
Suppose, we are given a pair of sensor data sequences $\{x_a, x_b\}$: $x_a=(\textsl{x}_{a_1},...,\textsl{x}_{a_{\tau'}})$ and $x_b=(\textsl{x}_{b_1},...,\textsl{x}_{b_{\tau''}})$, where $\tau'$ and $\tau''$ denote the time length of the signals and $\textsl{x}_{a_t} = [s_{t}^{1},...,s_{t}^n]$ is the $n$\madd{-}dimensional sensor reading in sequence $x_a$ at time $t$. The learned mappings are then defined as follows:
\begin{equation}
  f^{act}:\begin{cases}
    (\textsl{x}_{a_1},...,\textsl{x}_{a_{\tau'}}) \rightarrow H^{act}_{x_a} \\
    (\textsl{x}_{b_1},...,\textsl{x}_{b_{\tau''}}) \rightarrow H^{act}_{x_b}
  \end{cases}
\end{equation}

\begin{equation}
  f^{pers}:\begin{cases}
    (\textsl{x}_{a_1},...,\textsl{x}_{a_{\tau'}}) \rightarrow H^{pers}_{x_a} \\
    (\textsl{x}_{b_1},...,\textsl{x}_{b_{\tau''}}) \rightarrow H^{pers}_{x_b}
  \end{cases}
\end{equation}

\begin{figure}
\begin{center}
\includegraphics[width=0.55\textwidth]{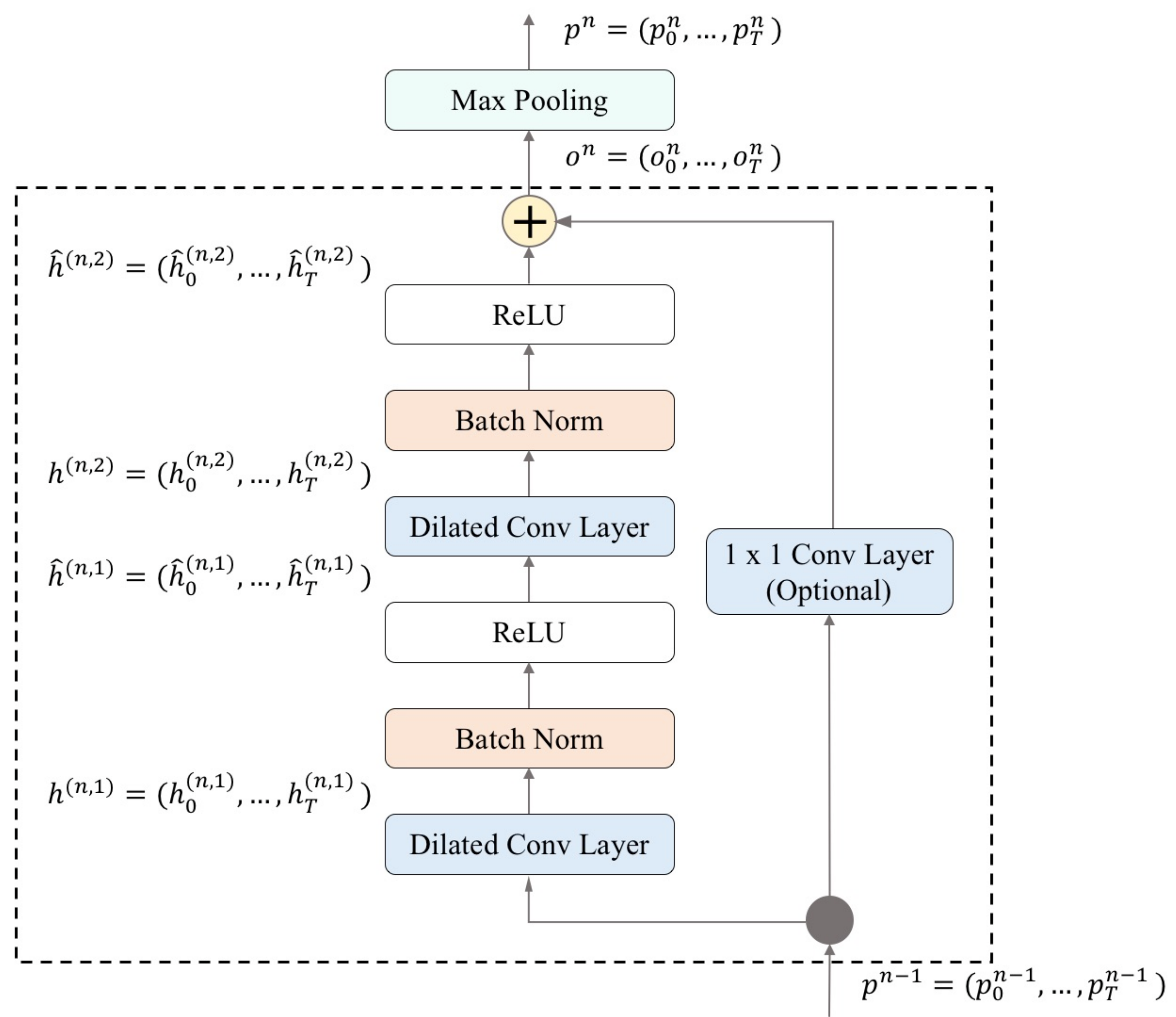}
\caption{A TCN block with two convolutional layers.} \label{tcn}
\com{Is this too small ?}
\end{center}
\end{figure}

The TCN architecture is adopted as the basic building block for abstracting the sequence due to its ability to efficiently abstract time series data at different time scales. As illustrated in Fig.~\ref{tcn}, each TCN block is composed of a series of transformations, which includes the dilated temporal convolutions with dilation rate $d$, batch normalization, non-linearity $g(\cdot)$, and residual connection $\oplus$. 
Assume that the proposed model contains a sequence of $N$ TCN blocks where each block contains $L$ convolutional layers with $m$ different feature maps of width $k$, and the parameters of the feature maps are $w$. For the $n$\textsuperscript{th} block, the input $p^{n-1}$ is the max-pooled output from the $(n-1)$\textsuperscript{th} block, \mdel{then}\madd{and} \com{Maybe this should be an if-then ?} the temporal convolutions that we appl\del{ied}\add{y} in each TCN block to capture the patterns over the course of an activity \del{is}\add{are} \del{defined}\add{constructed} \del{with}\add{from} \add{the} following \del{techniques}\add{components}.

\subsubsection{Dilated Convolutional Layer}The advantage of dilated convolutions is that \del{it}\add{they} support\del{s} faster expanding receptive fields without losing resolution or coverage \cite{MULTISCALCONtext}. Batch Normalization (BN) is applied after each dilated convolutional layer and before the non-linear activation function. BN can accelerate the learning process by reparameterizing the underlying optimization problem to make it more stable and smooth \cite{bnho}. Formally, in the $l^{th}$ convolutional layer of the TCN block, the computation is then defined as follows:
\begin{equation}
  h_{t}^{(n,l)} =\begin{cases}
    \sum_{i=-\frac{k-1}{2}}^{\frac{k-1}{2}}w_i \cdot p^{n-1}_{t-d \cdot i}, & \text{if $l=1$}.\\
    \sum_{i=-\frac{k-1}{2}}^{\frac{k-1}{2}}w_i \cdot \hat{h}_{t-d \cdot i}^{(n,l-1)}, & \text{otherwise}.
  \end{cases}
\end{equation}
\begin{align}
& \hat{h}_{t}^{(n,l)} = g(h_{t}^{(n,l)})
\end{align}
where $h^{(n,l)}$ is the output of the convolutions and $\hat{h}^{(n,l)}$ is the output of the non-linear activation function. 

\subsubsection{Residual Connections}The output of the TCN block, $o^n$, is the \madd{sum of the} result of the last convolution $\hat{h}_{t}^{(n,L)}$ \mdel{adding}\madd{and} the input of the block\madd{,} $p^{n-1}$. However, in TCN blocks, the shapes of the input tensor and the output tensor can be different \cite{cnnrnn}. To address this problem, if $\hat{h}_{t}^{(n,L)}$ and $p_{t}^{n-1}$ have different shapes, a 1x1 convolution will be used as the residual connection. Otherwise, an identity function will be used as the residual connection:
\begin{align}
& o_{t}^{n} = \hat{h}_{t}^{(n,L)} \oplus p_{t}^{n-1}
\end{align}

The temporal max pooling with width 2 is used between two consecutive TCN blocks to reduce the size of the time dimensions while introducing slight translational invariance in time:
\begin{align}
& p_{t}^n = max(o_{t-1}^n, o_{t}^n)
\end{align}

\subsection{Multi-Output Siamese Networks}
\begin{figure}
\begin{center}
\includegraphics[width=0.55\textwidth]{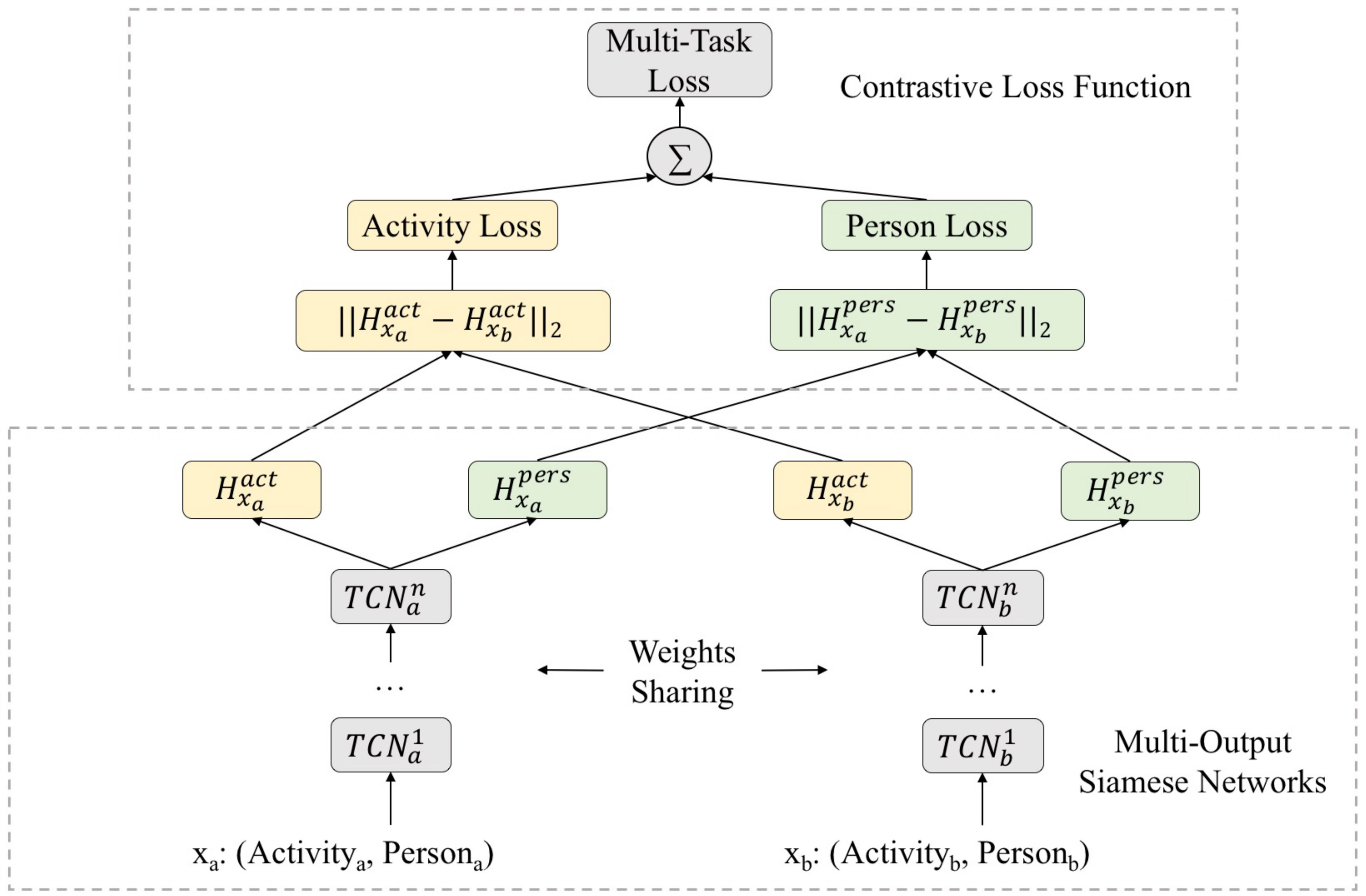} 
\caption{A dual-output siamese network\mmmdel{s}.} \label{siamese}
\com{Is this too small ?}
\end{center}
\end{figure}

The architecture of the proposed model is outlined in Fig.~\ref{siamese}. It contains two networks TCN\textsubscript{a} and TCN\textsubscript{b}. Each network has $n$ TCN blocks, shares the weights, and processes one of the data sequences in the input pair $\{x_a, x_b\}$. In order to disentangle the activity and person information extracted by the TCN networks, two fully-connected (FC) layers are connected to the TCN networks and are responsible to process the activity representations $\{H_{x_a}^{act}, H_{x_b}^{act}\}$ and person representations $\{H_{x_a}^{pers}, H_{x_b}^{pers}\}$, respectively. The weights of the FC layers are shared within the same representation learning task. 

This model is trained using 4-tuples $\{x_a, x_b, y^{act}, y^{pers}\}$, where $x_a$ and $x_b$ are the data sequences of the input pair. $y^{act}, y^{pers} \in \{0, 1\}$ denote the semantic relationships of the input pair, where $y^{act}=0$ or $y^{pers}=0$ denotes that $\{x_{a}, x_{b}\}$ is a semantically negative pair, i.e. they are maximally dissimilar in terms of the corresponding property. If $y^{act}=0$, $x_a$ and $x_b$ are different kinds of activities; if $y^{pers}=0$, $x_a$ and $x_b$ are performed by different persons. $y^{act}=1$ or $y^{pers}=1$ denotes that $\{x_a, x_b\}$ is a semantically positive pair. If $x_a$ and $x_b$ are the same kind of activity, $y^{act}=1$; if $x_a$ and $x_b$ are performed by the same person, $y^{pers}=1$.

\subsection{Contrastive Loss Function}
Given the input pair $\{x_a, x_b\}$, the model outputs a pair of activity representations\madd{,} $\{H^{act}_{x_a}, H^{act}_{x_b}\}$, and a pair of person representations $\{H^{pers}_{x_a}, H^{pers}_{x_b}\}$. The loss function used to train the model is also defined on the pairs. The similarity distance $Dist_s$ between the input pair is measured by the euclidean distance between their representation pair:
\begin{align}
& Dist_{s}(x_a, x_b) = ||H_{x_a}-H_{x_b}||_2
\end{align}
To make the notation clearer, $Dist_{s}(x_a, x_b)$ is rewritten as $D$. Then the loss function for each of the categorizations used for training is defined as:
\begin{align}
& L(x_a, x_b, y) = \sum_{i=1}^{N} L^i(x_{a}^{i}, x_{b}^{i}, y^i) \\
& L^i(x_{a}^{i}, x_{b}^{i}, y^i) = y^{i}L_{s}( x_{a}^{i}, x_{b}^{i})+(1-y^{i})L_{d}(x_{a}^{i}, x_{b}^{i}) 
\end{align}
where $(x_{a}^{i}, x_{b}^{i}, y^i)$ is the $i$-th sample in the data set. $L_s$, $L_d$ are the loss terms for the positive pair $(y=1)$ and the negative pair $(y=0)$. The forms of $L_s$, $L_d$ are given by:
\begin{align}
&L_s = \frac{1}{2}(D)^2 \\
&L_d=\frac{1}{2}\{max(0, \delta-D)\}^2
\end{align}
where $\delta$ is a margin hyperparameter. It defines that the negative pairs contribute to the loss function\mdel{,} if their distance is smaller than the margin $\delta$. 
Then, the loss function is applied in each representation space, and the weighted sum of the loss function in each representation space is defined for the final multi-task model ($\alpha$ and $\beta$ are the corresponding weights for each task):
\begin{align}
\begin{split}
&\mathcal{L}(x_a, x_b, y^{act}, y^{pers})=\\
&\alpha \cdot L^{act}(x_a, x_b, y^{act})+ \beta \cdot L^{pers}(x_a, x_b, y^{pers})
\end{split}
\end{align}
To train the network, we use the standard backpropagation algorithm with stochastic gradient descent. All the parameters are initialized to small values. The start learning rate is $lr=0.05$ and an exponential decay function is applied to the learning rate every 10000 steps with a decay rate of 0.95. The network is tested on validation data after each epoch. If the validation error stopped decreasing for a predefined number of epoch\mmmadd{s}, training is finished.

\subsection{Cluster\eqdel{s} Construction}
\label{cluster construction}
After all the parameters are learned, we can use the trained model to provide metrics for a wide range of different clustering algorithms. Because the trained model can map the data sample $x$ into the representation space, where the representation vector $H$ \del{are}\add{is} positioned \del{in the space} such that \del{the} data with the same semantic meaning are \del{closely} located \add{close} to each other\madd{, clusters in this space should capture the corresponding property}. More specifically, in our experiments, the trained model maps the data samples into the activity representation space and the person representation space. In the activity representation space, the data samples will be grouped into clusters according to the activity type. In the person representation space, the data samples will be grouped into clusters according to the identity of the person. Hence, after mapping the data samples to the more clustering-friendly representations, different clustering algorithms can be used on these learned representations. In our experiments, K-means is employed as the clustering method. 

\section{Evaluation and Experiments}
In order to evaluate the effectiveness of the proposed model, we use four public datasets that contain raw sensor data sequences of different human activities performed by different persons. We conduct activity clustering and person clustering on the learned representations as described in \mdel{s}\madd{S}ection \ref{cluster construction}.

\subsection{Datasets}
The datasets used here are selected from \mdel{those} widely used benchmark datasets \cite{Wang2018DeepLF}\mdel{, which should} \madd{as the ones that} contain a good number of different persons performing numerous diverse activities. Those datasets are recorded by various sensors, e.g. accelerometer, gyroscope, magnetometer etc, and include human activities in different scenarios. All the sensor data sequences are segmented with a sliding window as described with each of the datasets below.

The \textbf{PAMAP2} dataset is collected from 9 participants performing 12 activities over a total of 10 hours. It includes sport exercises (rope jumping, nordic walking etc), and household activities (vacuum cleaning, ironing etc). One heart-rate monitor and three inertial measurement units (IMUs) located on the chest, dominant wrist and ankle were used to record the heart rate, accelerometer, gyroscope, magnetometer, and temperature data. We replicate previous work \cite{understandRNNHAR,dnnHARbenchmark} to downsample the data from 100Hz to 33.3Hz, and use a sliding window of 5.12 seconds with one second step size.  % resulting a training-set with around. 473k samples.

The \textbf{MHEALTH} dataset contains data recorded from 10 volunteers carrying out 12 physical activities, including primitive body parts movements (waist bends forward, frontal elevation of arms etc), and composite body movements (cycling, jumping front and back etc). The data is collected by using three sensors placed on the subject's chest, right wrist and left ankle to record accelerometer, gyroscope, and magnetometer signals. The chest sensor also records 2-lead ECG signals. The sampling rate of all sensing modalities is 50Hz. As in previous work \cite{limited}, we use a sliding window of 5 seconds with a step size of 2.5 seconds.

The \textbf{SBHAR} dataset provides data gathered from 30 participants performing 6 basic activities, such as walking and lying, and 6 postural transitions, such as stand-to-sit, sit-to-lie. In our experiment, we consider all the postural transitions as one general transition. The data was collected by placing a smartphone on the waist of the participant, and the inertial sensors in the smartphone were used to record the accelerometer and gyroscope data at a sampling rate of 50Hz. As used in the previous work \cite{transitionAware}, we use a sliding window of 2.56 seconds with a step size of 1.28 seconds.

The \textbf{WISDM} dataset contains data collected with one accelerometer in a smartphone from 36 volunteers carrying out 6 activities, including jogging, climbing stairs, etc. The sampling rate is 20Hz. We use the same settings as used in \cite{wisdm} to set the sliding window size to be 10 seconds without overlap.

\subsection{Performance Metrics}
To compare with previous works, we use mean F1-score ($F_m$) and clustering accuracy as the metrics. $F_m$ is defined as follows:
\begin{align}
&F_m = \frac{2}{|C|} \sum_{i=1}^{C} \frac{Precision_i \cdot Recall_i}{Precision_i + Recall_i}  %\\
%&F_w = 2\sum_{i=1}^{C} \frac{N_i}{N_{total}} \frac{Precision_i \cdot Recall_i}{Precision_i + Recall_i} 
\end{align}
where $i = 1, \ldots, C$ is the set of classes. For the given class $i$, $Precision_i = \frac{TP_i}{TP_i + FP_i}$, $Recall_i = \frac{TP_i}{TP_i + FN_i}$; $TP_i$ and $FP_i$ denote the number of true positives and false positives, and $FN_i$ is the number of false negatives. %$N_i$ is the number of samples in class $i$, $N_{total}$ is the total number of samples.

\subsection{Results}
\label{Results}
\begin{table}
\centering
\caption{Results on PAMAP2 in terms of $F_m$}\label{pamap2}
%\vspace*{-0.15in}
\scalebox{0.85}{%
\begin{tabular}{|c|c|c|}
\hline
\textbf{Methods} &  \textbf{Activity} & \textbf{Person} \\
\hline
Probability SVM \cite{transitionAware} &  0.9304 & -\\
Probability SVM with Filter \cite{transitionAware} &  0.9433 & -\\
LSTM-F \cite{dnnHARbenchmark} & 0.9290 & -\\
CNN \cite{dnnHARbenchmark} & 0.9370 & -\\
DNN \cite{dnnHARbenchmark} & 0.9040 & -\\
%Rotation Forest \cite{medpamap2} & 0.9801 & -\\
\hil{State + Person SVMs \cite{piPositionKinect}} & \hil{0.9384} & \hil{0.9123} \\
PCA + k-Means & 0.4244 & 0.2040\\
3-NN & - & 0.9416\\
5-NN & - & 0.9014\\
Decision Tree & - & 0.9781\\
Proposed Multi-Task Method & \textbf{0.9893} & \textbf{0.9881}\\
\hline
\end{tabular}}
\end{table}
\begin{table}
\centering
\caption{Results on MHEALTH in terms of Accuracy}\label{mhealth}
%\vspace*{-0.15in}
\scalebox{0.85}{%
\begin{tabular}{|c|c|c|}
\hline
\textbf{Methods} &  \textbf{Activity} & \textbf{Person}\\
\hline
CNN-1D \cite{cnn2d}  & 0.9809 & -\\
CNN-2D \cite{cnn2d}  & 0.9829 & -\\
CNN-pff \cite{cnnpff}  & 0.9194 & -\\
\hil{FE-AT \cite{limited} } & \hil{0.9664} & -\\
\hil{State + Person SVMs \cite{piPositionKinect}} & \hil{0.8376} & \hil{0.6777} \\
PCA + k-Means & 0.4850 & 0.2361\\
3-NN & - & 0.8540\\
5-NN & - & 0.8380\\
Decision Tree & - & 0.6714\\
Proposed Multi-Task Method & \textbf{0.9957} & \textbf{0.9948}\\
\hline
\end{tabular}}
\end{table}
\begin{table}
\centering
\caption{Results on SBHAR in terms of Accuracy}\label{sbhar}
%\vspace*{-0.15in}
\scalebox{0.85}{%
\begin{tabular}{|c|c|c|}
\hline
\textbf{Methods} &  \textbf{Activity} & \textbf{Person} \\
\hline
Probability SVM \cite{transitionAware}& 0.9580 & -\\
Probability SVM with Filter \cite{transitionAware}  & 0.9678 & -\\
CNN \cite{lundHAR}&  0.9870 & -\\
%ERNN \cite{ERNN}&  0.9633 & -\\
\hil{State + Person SVMs \cite{piPositionKinect}} & \hil{0.9034} & \hil{0.5614} \\
PCA + k-Means & 0.5282 & 0.1299\\
3-NN & - & 0.5330\\
5-NN & - & 0.5123\\
Decision Tree & - & 0.4957\\
Proposed Multi-Task Method & \textbf{0.9885} & \textbf{0.8892}\\
\hline
\end{tabular}}
\end{table}
\begin{table}
\caption{Results on WISDM in terms of Accuracy}\label{wisdm}
%\vspace*{-0.15in}
\scalebox{0.85}{%
\centering
\begin{tabular}{|c|c|c|}
\hline
\textbf{Methods} &  \textbf{Activity}& \textbf{Person}\\
\hline
CNN with partial weight sharing \cite{cnnHAR}  &  0.9688 & -\\
DBN+HMM \cite{DBNHAR} &  \textbf{0.9823} & -\\
\hil{Decision Tree \cite{wisdm}} &  \hil{0.8510} & -\\
\hil{Multilayer Perceptron \cite{wisdm}} &  \hil{0.9170} & -\\
CNN+stat. features \cite{cnnstat} &  0.9332 & -\\
\hil{State + Person SVMs \cite{piPositionKinect}} & \hil{0.8128} & \hil{0.5070} \\
PCA + k-Means & 0.5181 & 0.2430\\
Neural Net$^{*}$ \cite{cbbi} &- & 0.6950\\
Decision Tree$^{*}$ \cite{cbbi} &- & 0.7220\\
3-NN & - & 0.2651\\
5-NN & - & 0.2470\\
Decision Tree & - & 0.2189\\
Proposed Multi-Task Method &0.9576 & \textbf{0.8112}\\
\hline
\end{tabular}}
\begin{tablenotes} 
\footnotesize
\centering
\item $^{*}$ Experiment results based on 4 activities.
\end{tablenotes}
\end{table}
We used the same model architecture \madd{with $3$ TCN blocks} across all the experiments. A shorthand description of the shared layers is: TCN $(128)-P-$ TCN $(128)-P-$ TCN $(128)$, where TCN$(128)$ denotes a TCN block with 128 feature maps, and $P$ a max-pooling layer. The internal structure of a TCN block is the same as illustrated in Figure~\ref{tcn}. \mdel{Moreover}\madd{In addition}, above the last TCN block, one FC layer with 256 hidden nodes is used for each single representation learning task. The experimental results are summarized in Table\add{s}~\ref{pamap2}, ~\ref{mhealth}, ~\ref{sbhar}, and ~\ref{wisdm}.

In addition to previous published works \madd{which use fully supervised learning}, we also used principal components analysis (PCA), k-nearest neighbors ($k=3, 5$), and a decision tree algorithm (C4.5) as baselines. As shown in the result tables, our proposed multi-task method achieved competitive performance on both tasks compared with \mdel{other}\madd{the} supervised single-task approaches in most situations. \madd{Since most previous methods were applied only to one of the tasks, only the performance results for that task are listed in the tables.}
\begin{figure*}
   \centering
    \begin{subfigure}{0.3\linewidth}        %% or \columnwidth
        \includegraphics[width=1.13\linewidth]{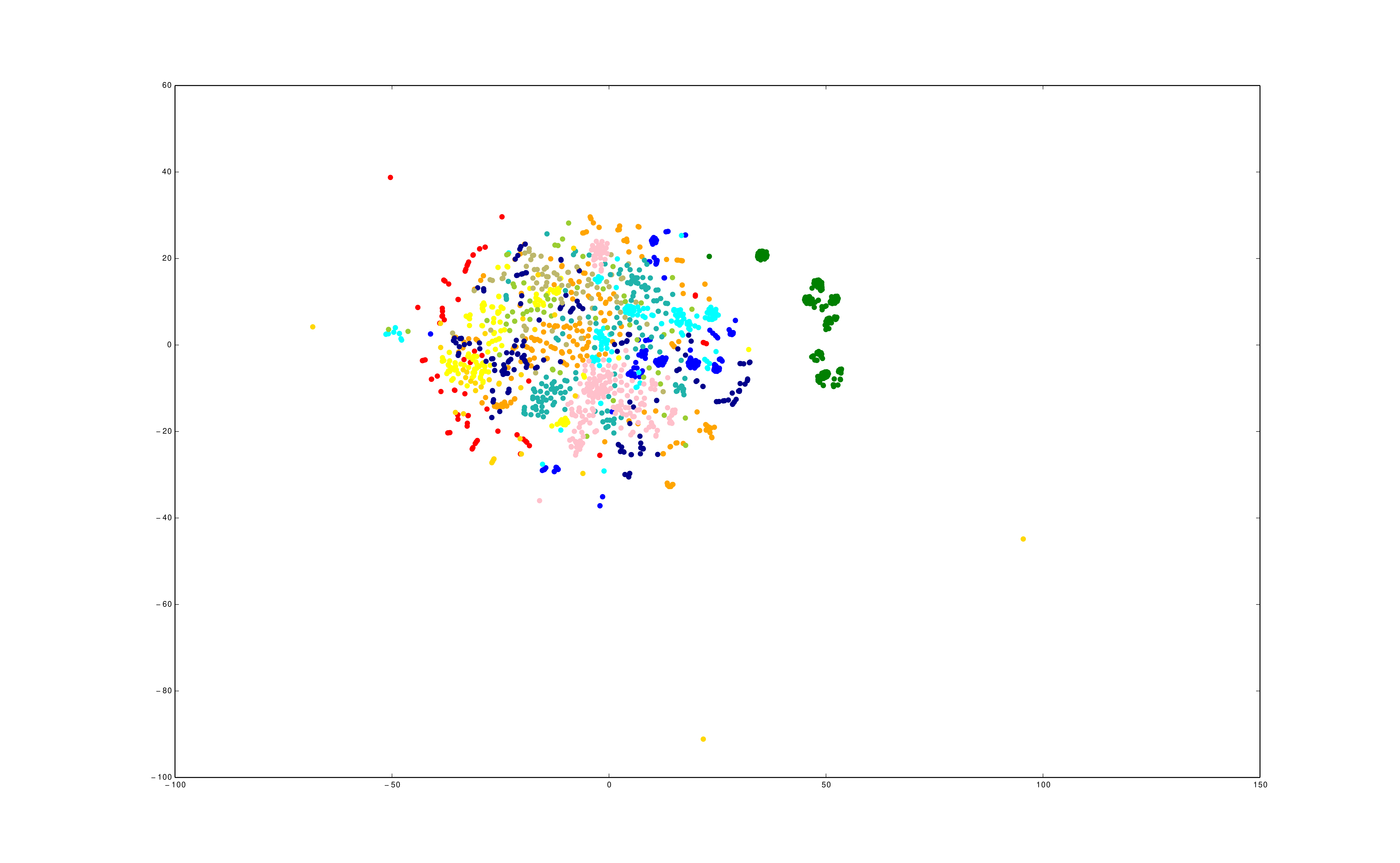}
        \caption{Original input space}
        \label{PAMAP2actOrigi}
    \end{subfigure}
    \begin{subfigure}{0.3\linewidth}        %% or \columnwidth
        \includegraphics[width=1.13\linewidth]{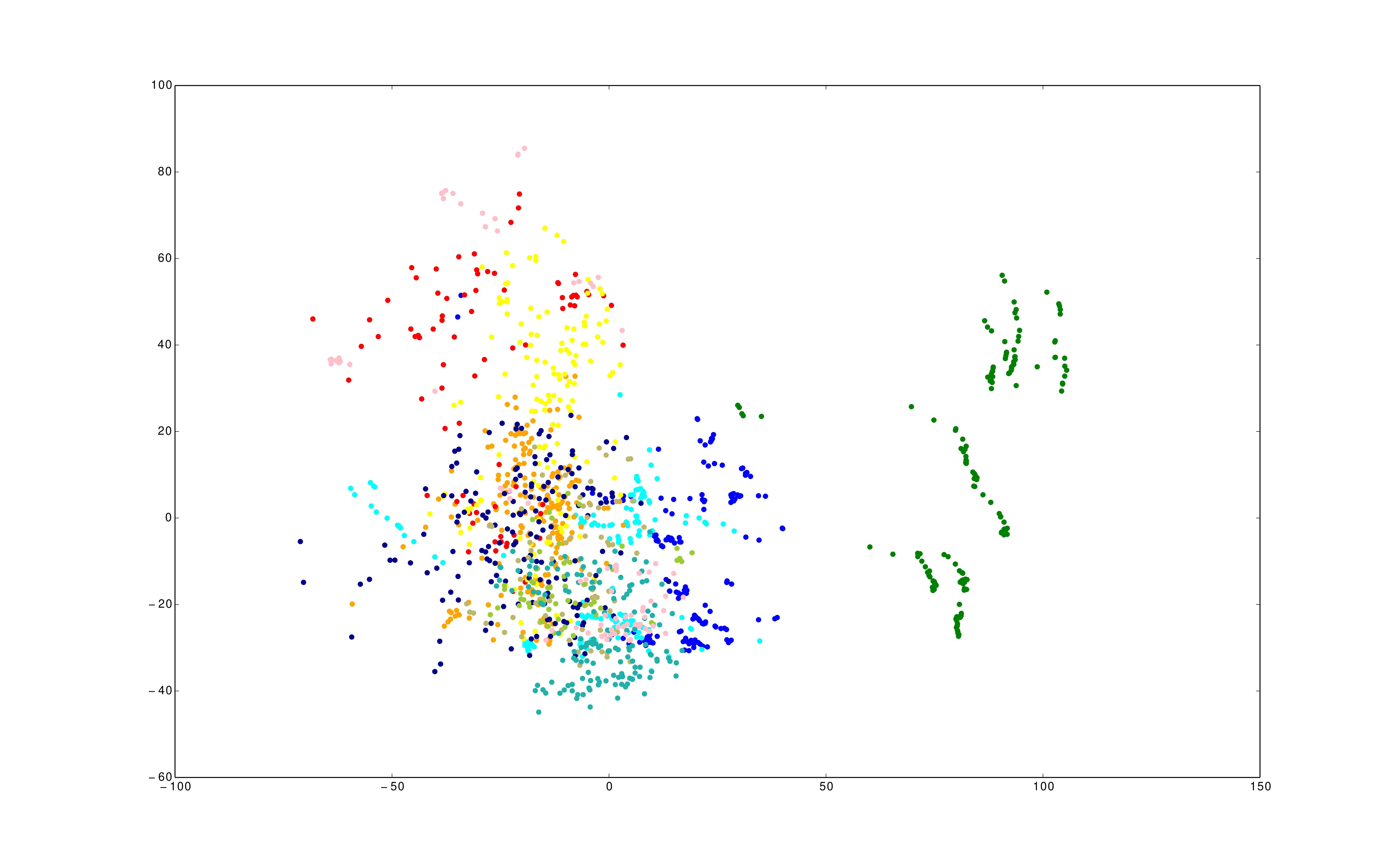}
        \caption{PCA}
        \label{PAMAP2actPCA}
    \end{subfigure}
    \begin{subfigure}{0.3\linewidth}        %% or \columnwidth
        \includegraphics[width=1.13\linewidth]{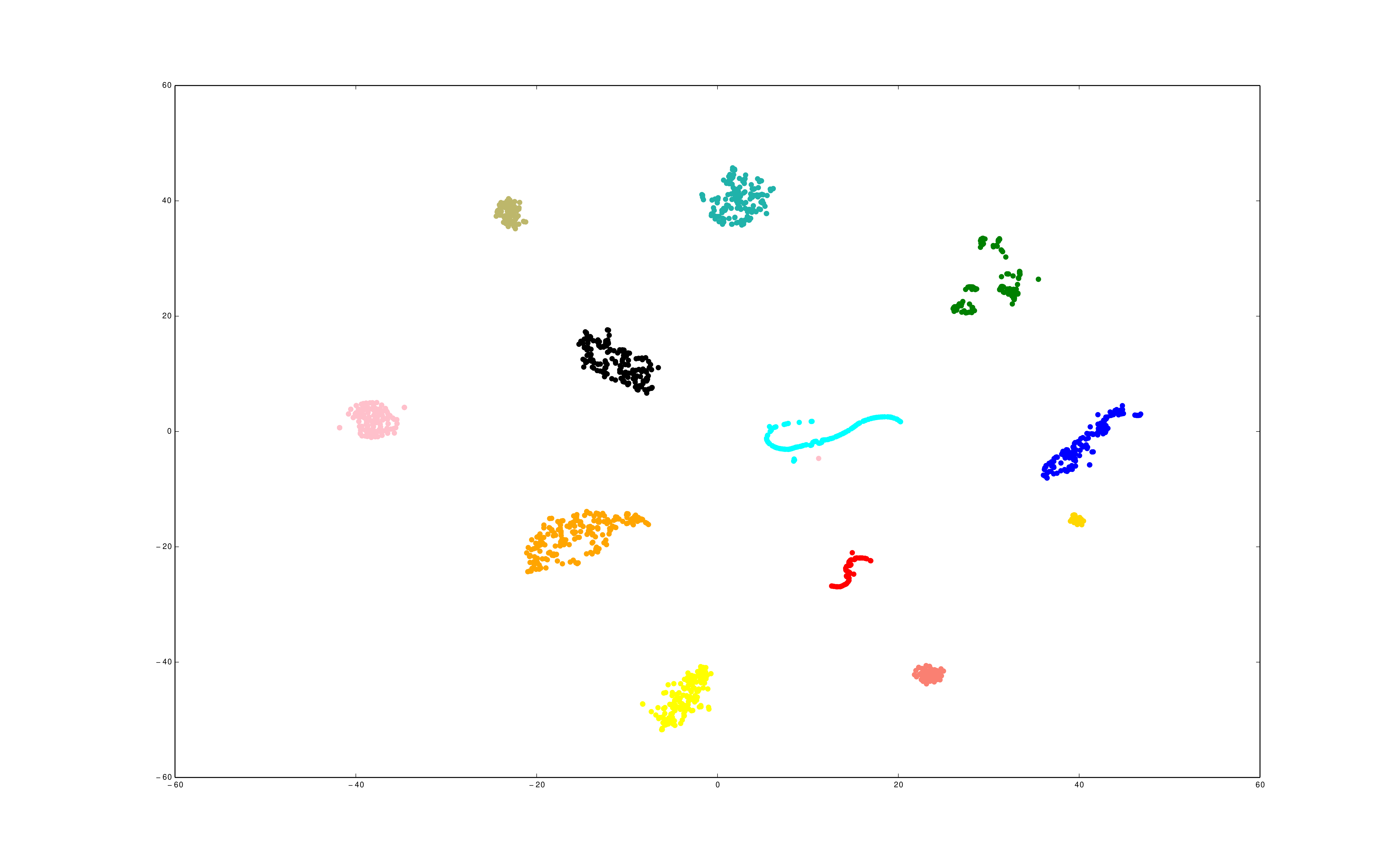}
        \caption{Proposed multi-task method}
        \label{PAMAP2actPro}
    \end{subfigure}   
    %\vspace*{-0.1in}
    \caption{Visualizations on the activity aspect of PAMAP2.}\label{PAMAP2act}
\end{figure*}
\begin{figure*}
    \centering    
    \begin{subfigure}{0.3\linewidth}        %% or \columnwidth
        \includegraphics[width=1.13\linewidth]{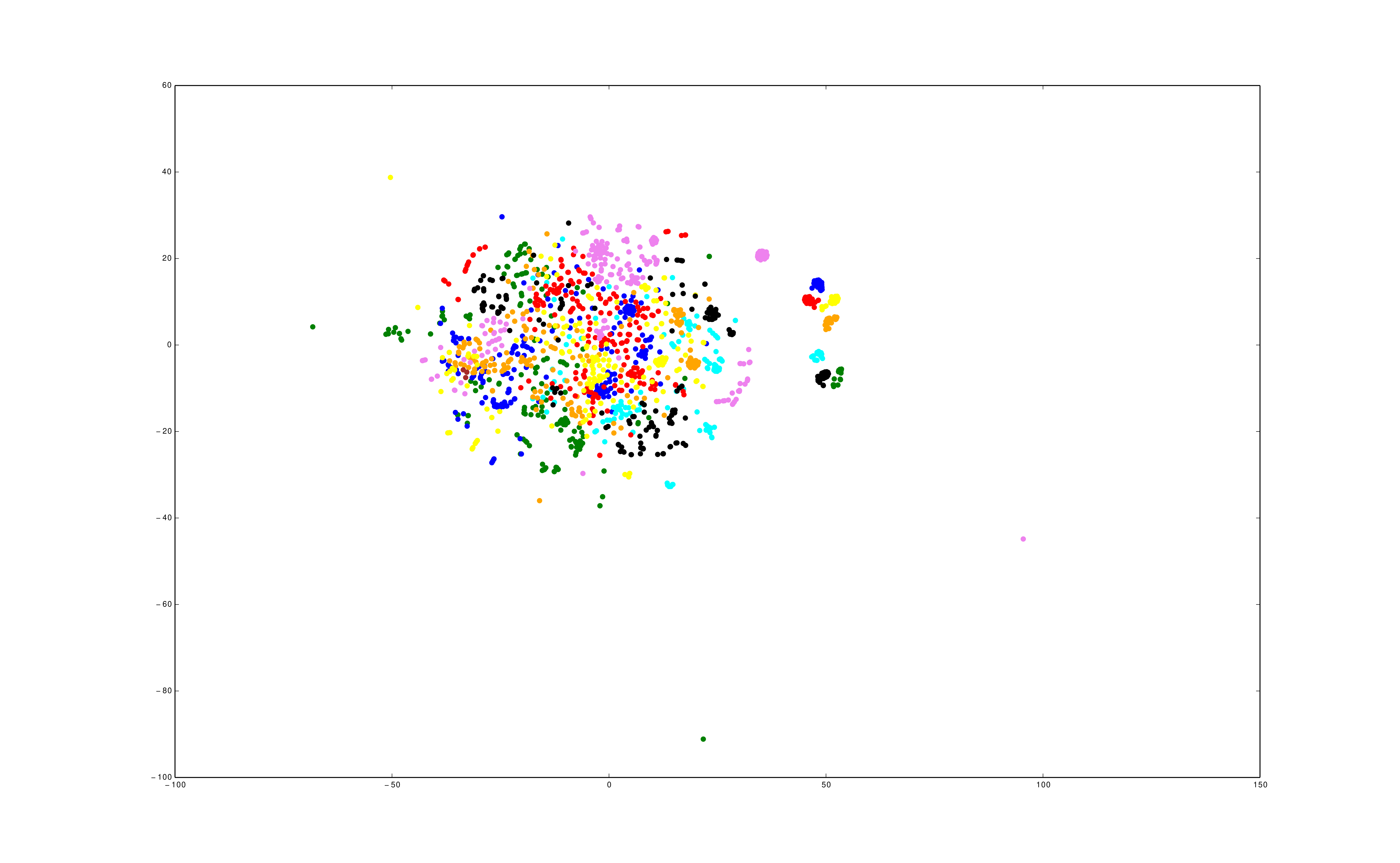}
        \caption{Original input space}
        \label{PAMAP2persOrigi}
    \end{subfigure}
    \begin{subfigure}{0.3\linewidth}        %% or \columnwidth
        \includegraphics[width=1.13\linewidth]{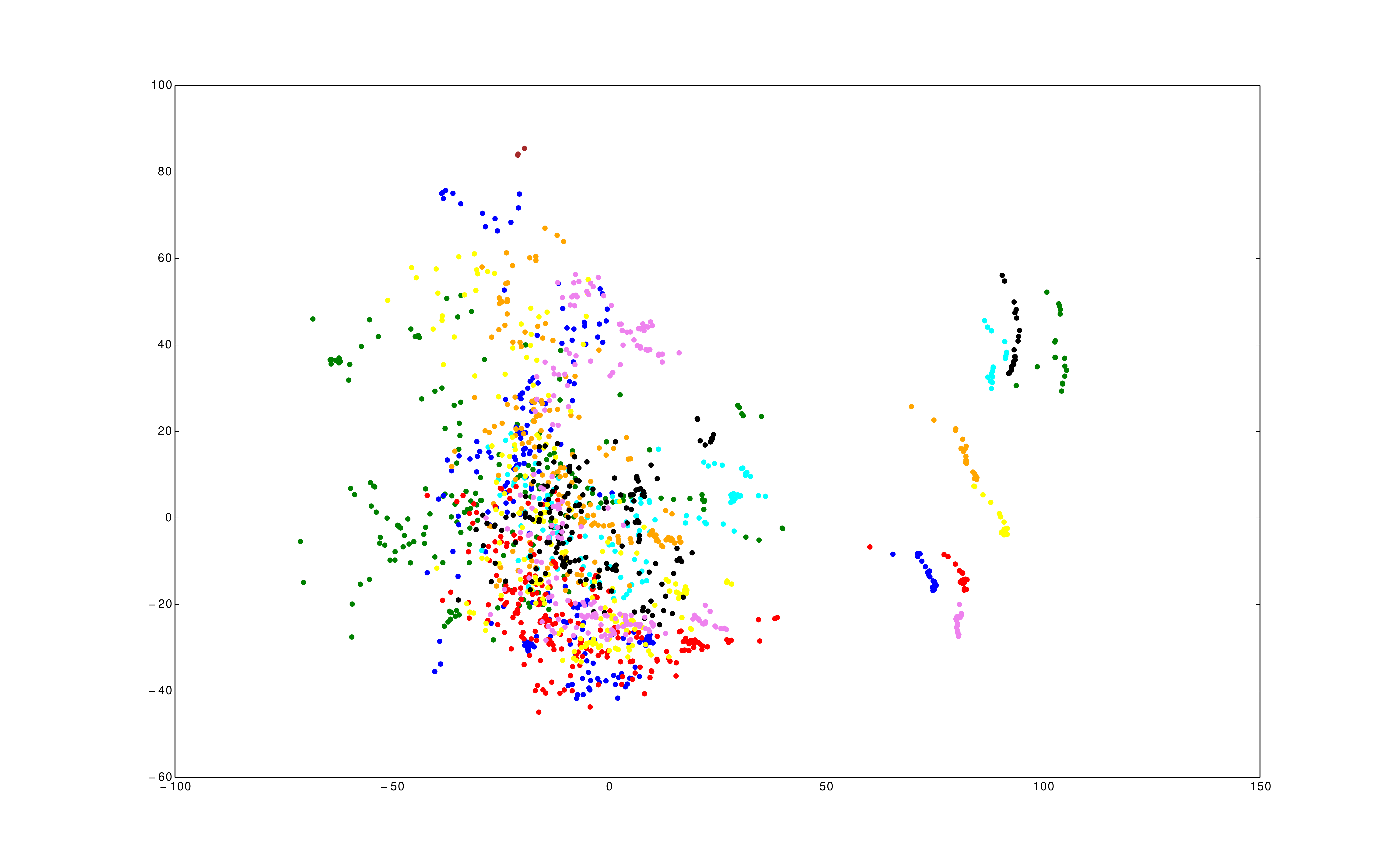}
        \caption{PCA}
        \label{PAMAP2persPCA}
    \end{subfigure}    
    \begin{subfigure}{0.3\linewidth}        %% or \columnwidth
        \includegraphics[width=1.13\linewidth]{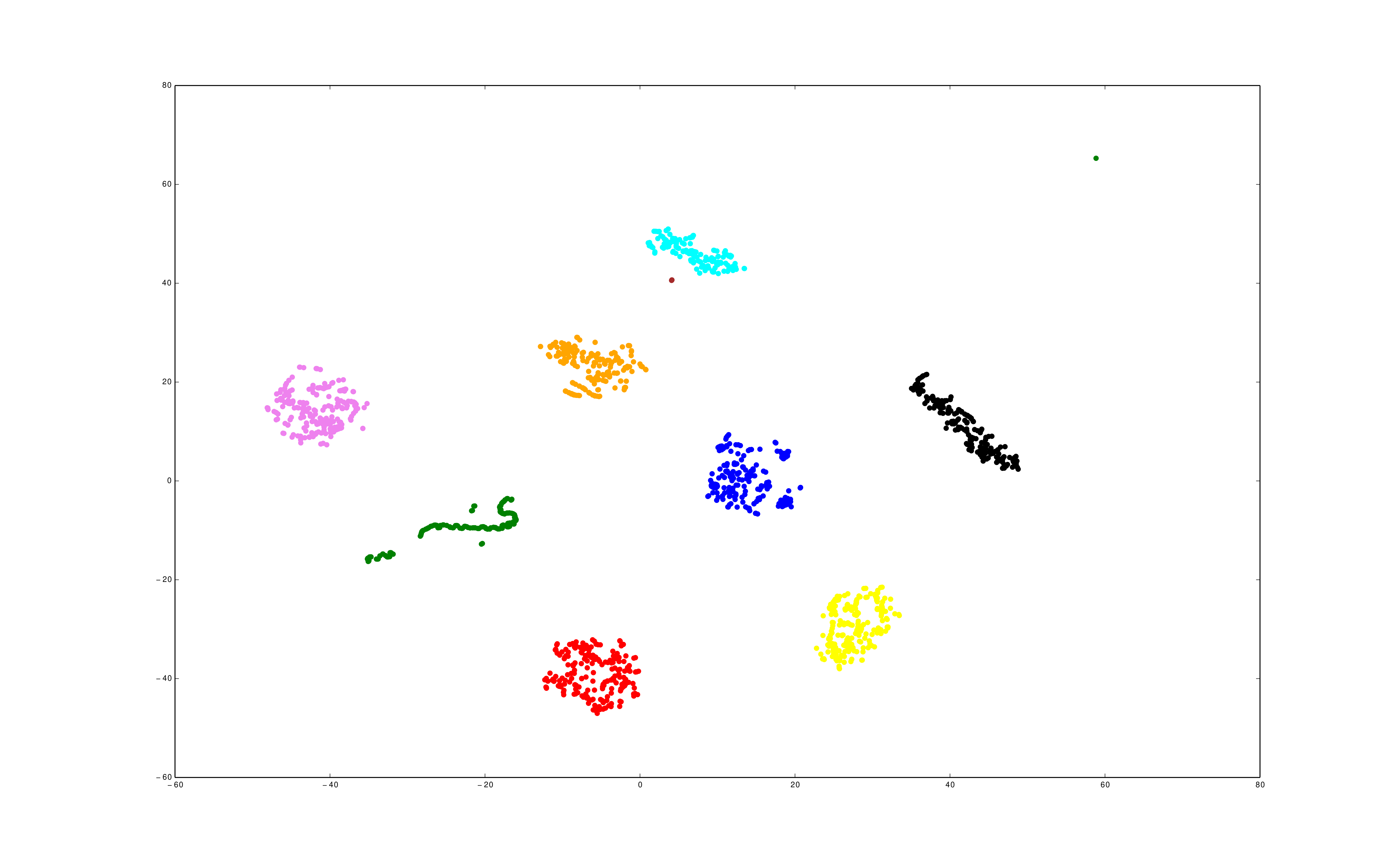}
        \caption{Proposed multi-task method}
        \label{PAMAP2persPro}
    \end{subfigure}
    %\vspace*{-0.1in}
    \caption{Visualizations on the person aspect of PAMAP2.}\label{PAMAP2pers}
\end{figure*}

\subsection{Visualization and Analysis}
To further analyze the representations learned by the proposed multi-task method and compare it with other embedding techniques, we used t-sne \cite{tsne}\com{Is there a good reference to include here ?} to visualize the activity and person representation spaces. Due to the space limitation, we only list the visualizations of the proposed multi-task method and PCA on two representative datasets: PAMAP2 and SBHAR. PAMAP2 contains complex activities, which makes it difficult to model. SBHAR contains unbalanced class numbers in the two aspects of the data. In particular, it includes 30 persons who perform 7 activities, which leads to unbalanced sample sets across the two tasks in that there are significantly fewer data samples for the person classes than for the activity classes. This, in turn, leads to bias and difficulties learning these two highly unbalanced tasks simultaneously. 

\begin{figure*}
    \centering
    \begin{subfigure}{0.3\linewidth}        %% or \columnwidth
        \includegraphics[width=1.13\linewidth]{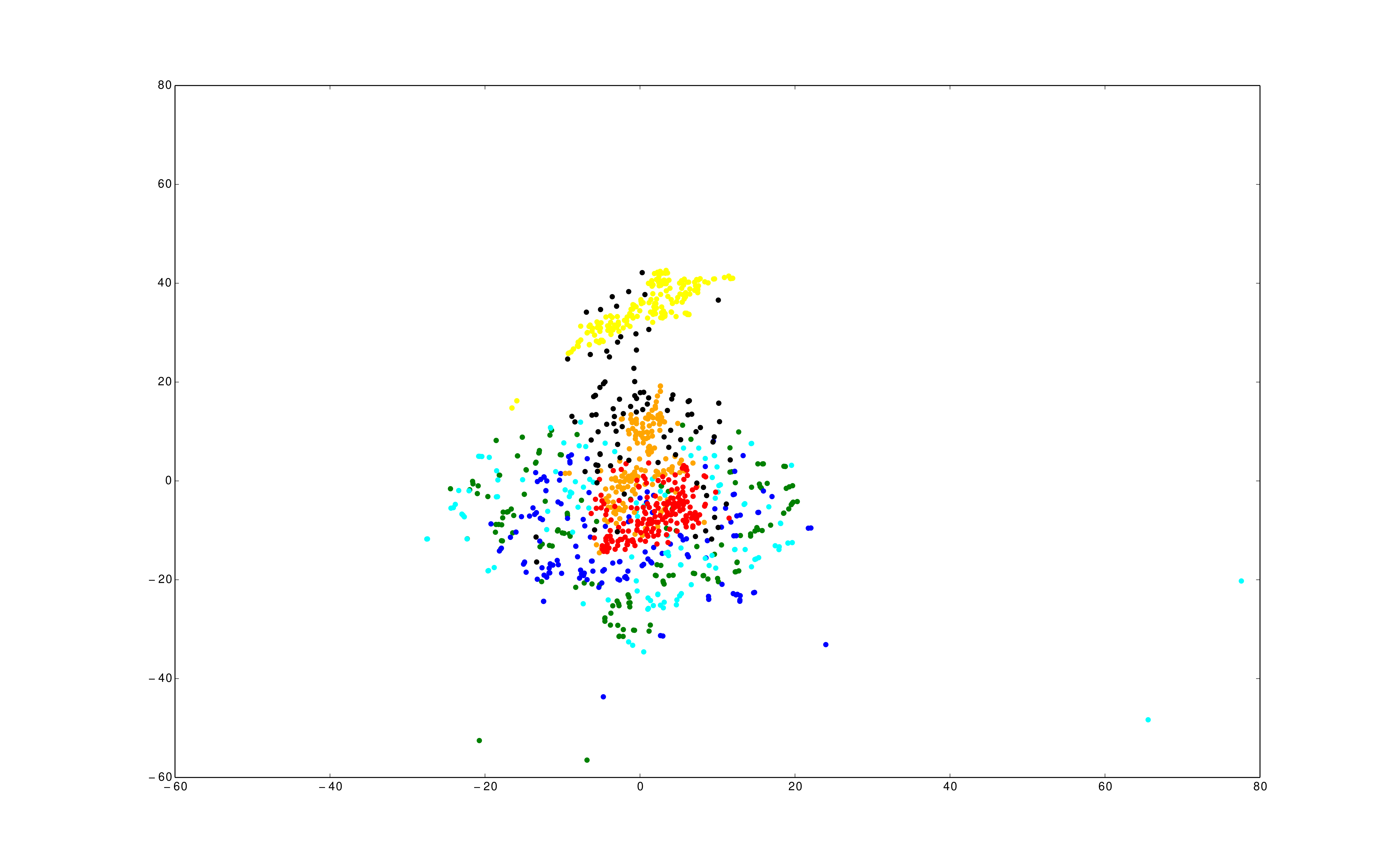}
        \caption{Original input space}
        \label{SBHARactOrigi}
    \end{subfigure}
    \begin{subfigure}{0.3\linewidth}        %% or \columnwidth
        \includegraphics[width=1.13\linewidth]{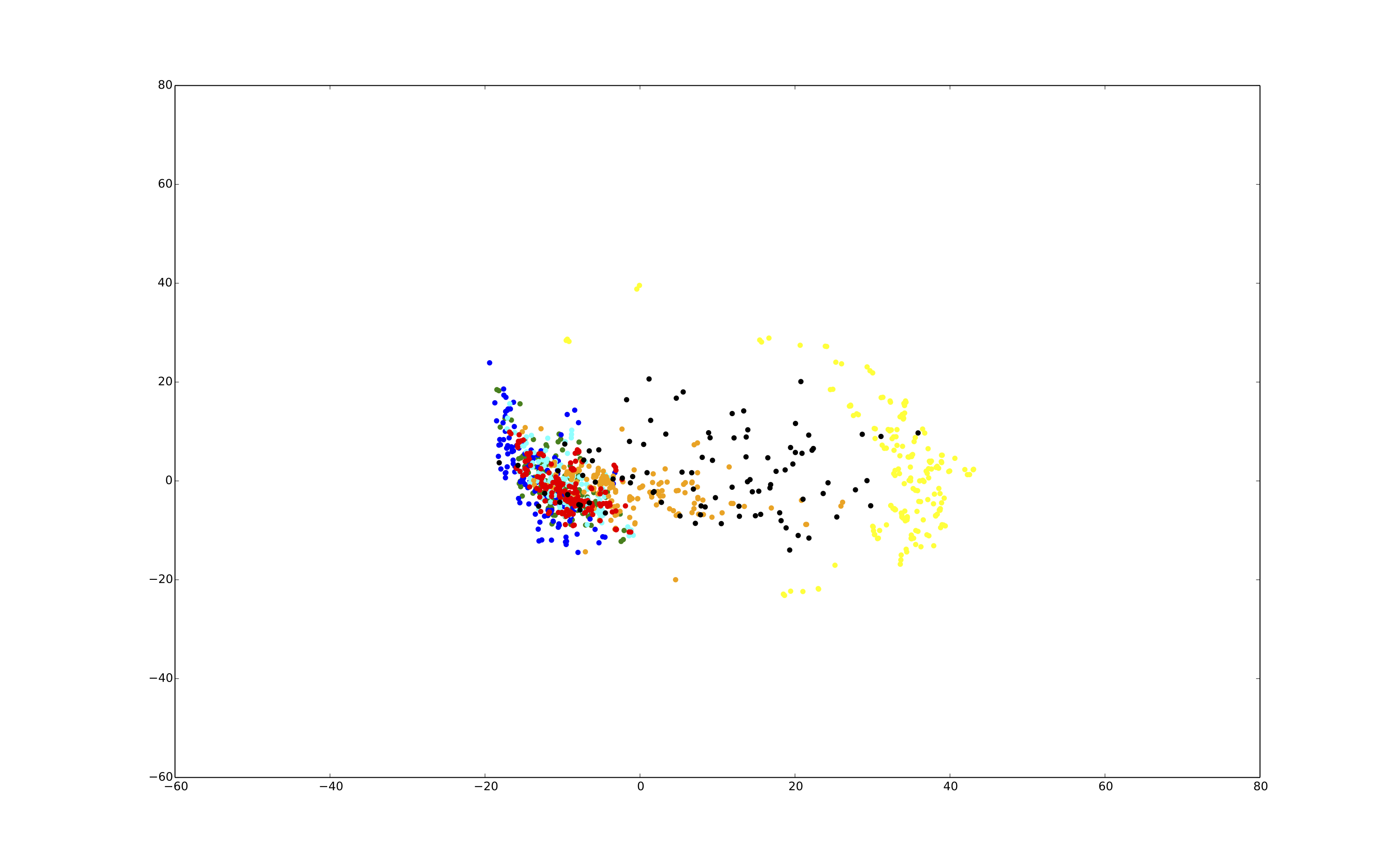}
        \caption{PCA}
        \label{SBHARactPCA}
    \end{subfigure}
    \begin{subfigure}{0.3\linewidth}        %% or \columnwidth
        \includegraphics[width=1.13\linewidth]{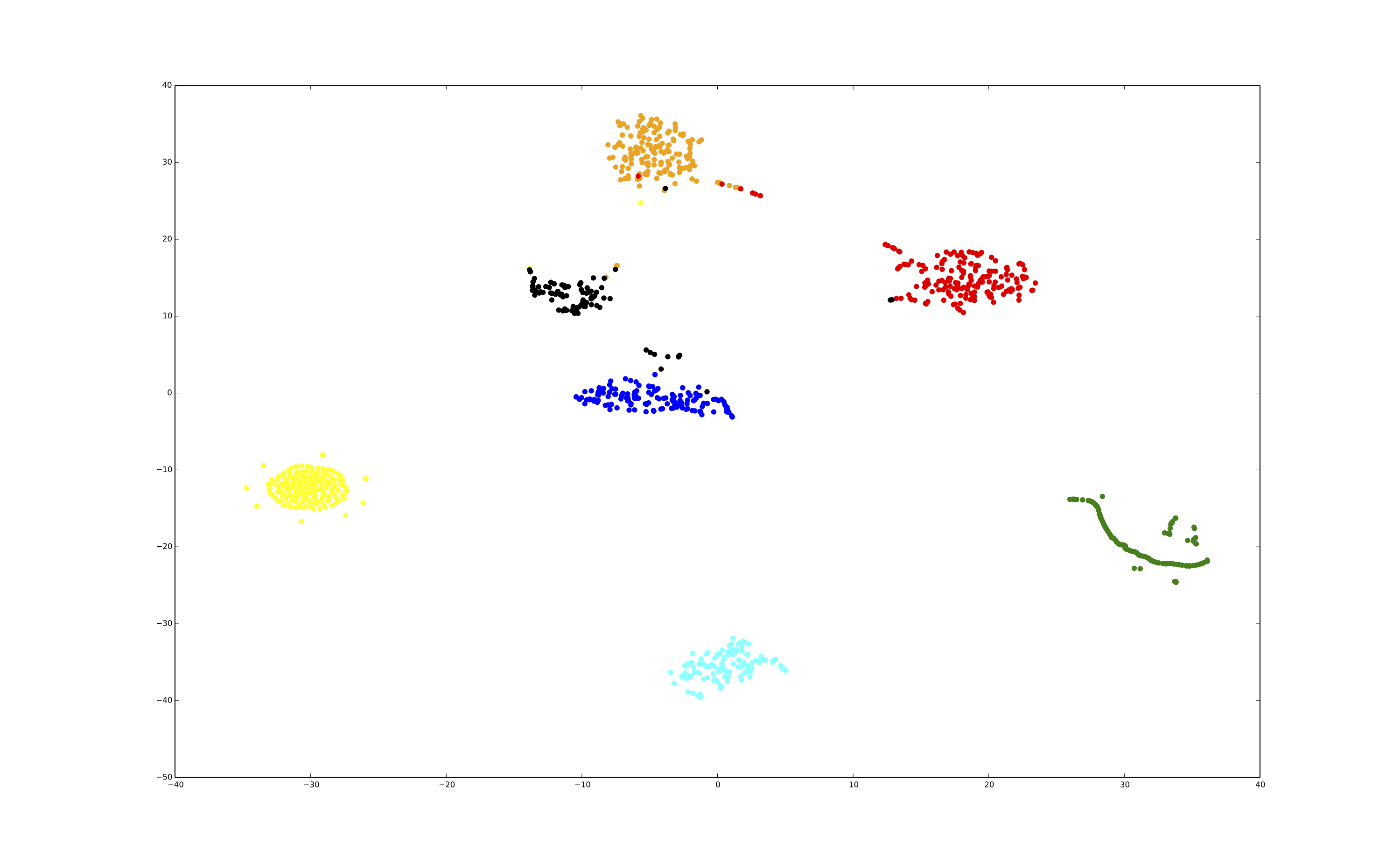}
        \caption{Proposed multi-task method}
        \label{SBHARactPro}
    \end{subfigure}
    %\vspace*{-0.1in}
    \caption{Visualizations on the activity aspect of SBHAR.}\label{SBHARact}
\end{figure*}

\begin{figure*}
    \centering
    \begin{subfigure}{0.3\linewidth}        %% or \columnwidth
        \includegraphics[width=1.13\linewidth]{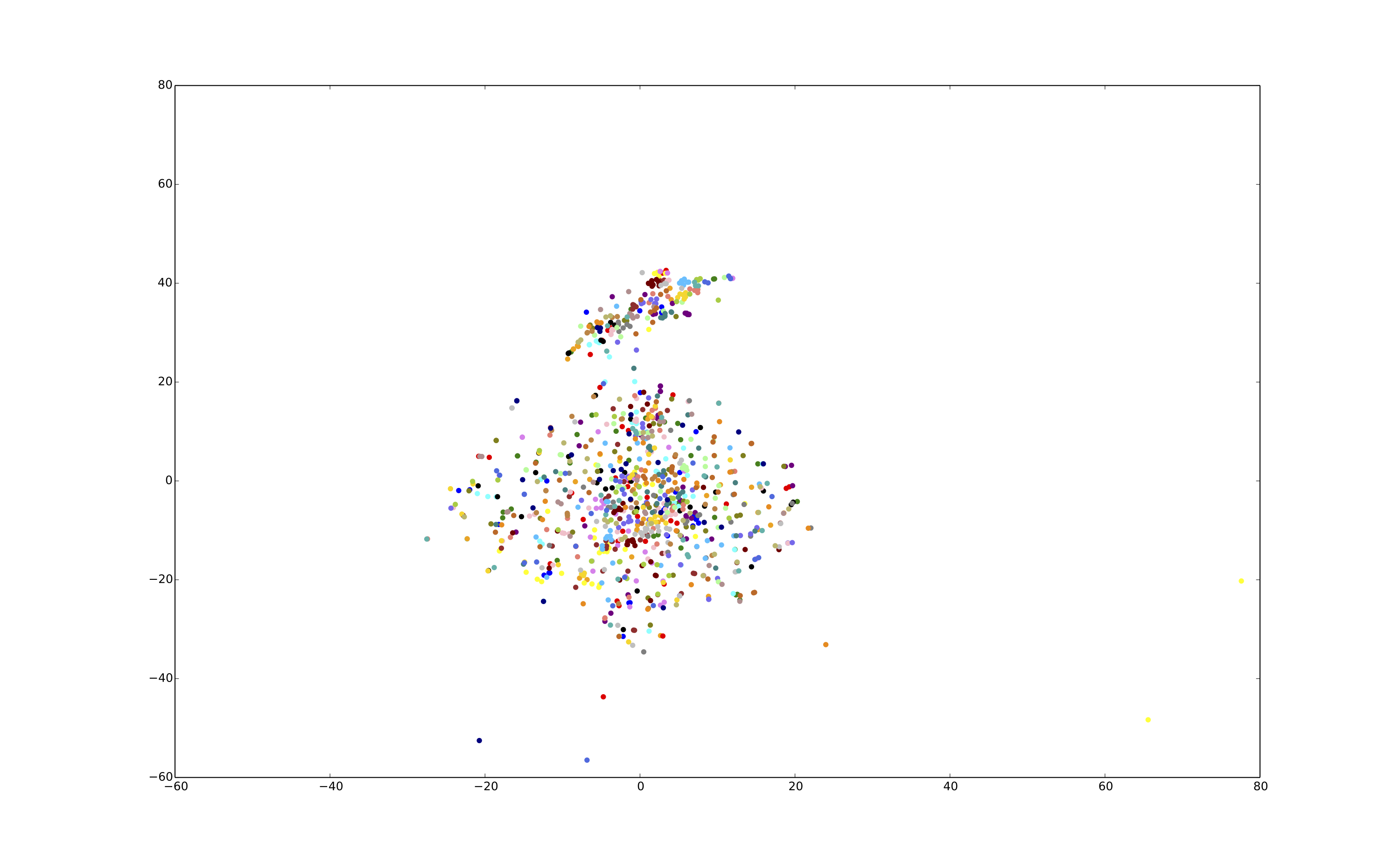}
        \caption{Original input space}
        \label{SBHARpersOrigi}
    \end{subfigure}
    \begin{subfigure}{0.3\linewidth}        %% or \columnwidth
        \includegraphics[width=1.13\linewidth]{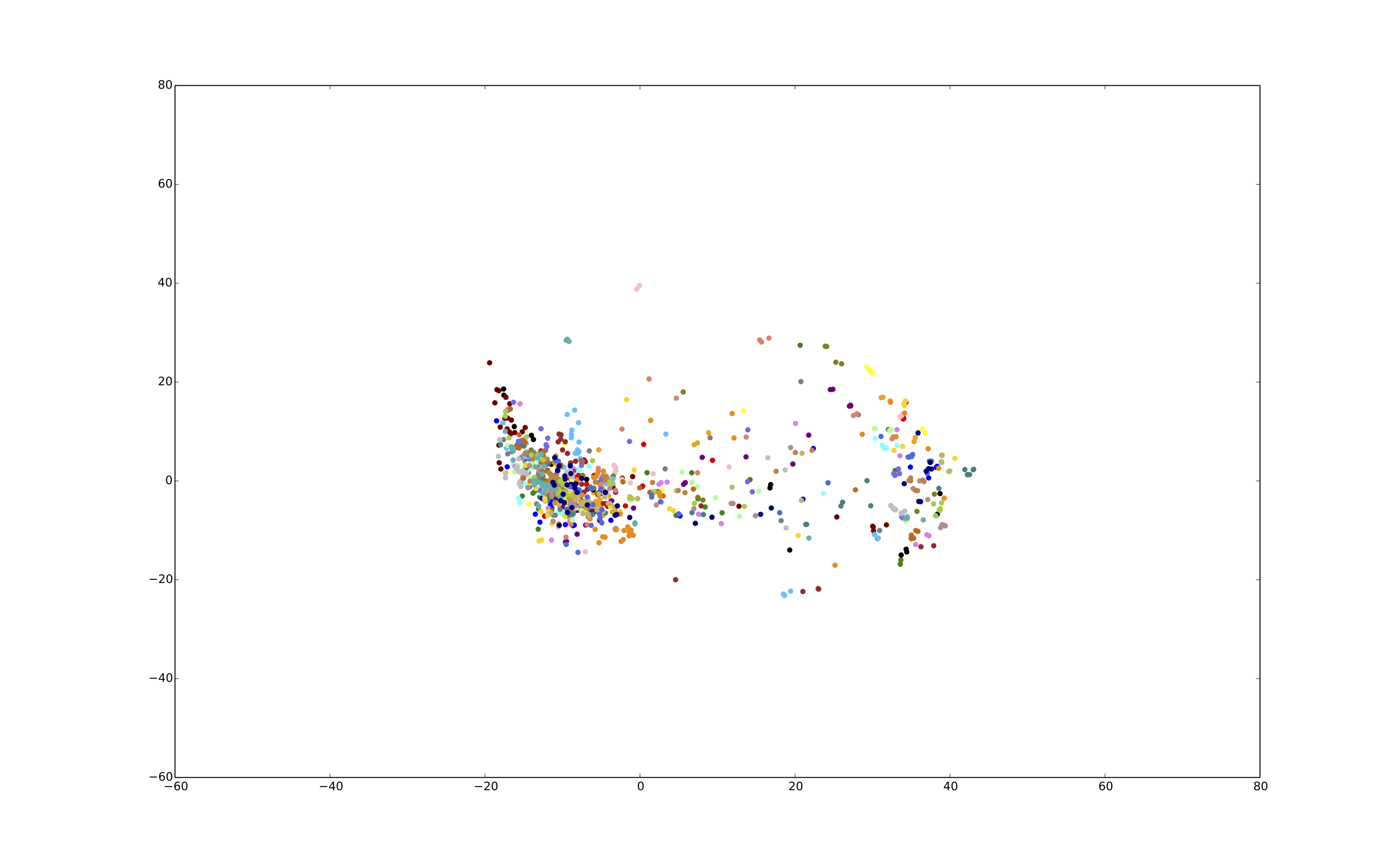}
        \caption{PCA}
        \label{SBHARpersPCA}
    \end{subfigure}
    \begin{subfigure}{0.3\linewidth}        %% or \columnwidth
        \includegraphics[width=1.13\linewidth]{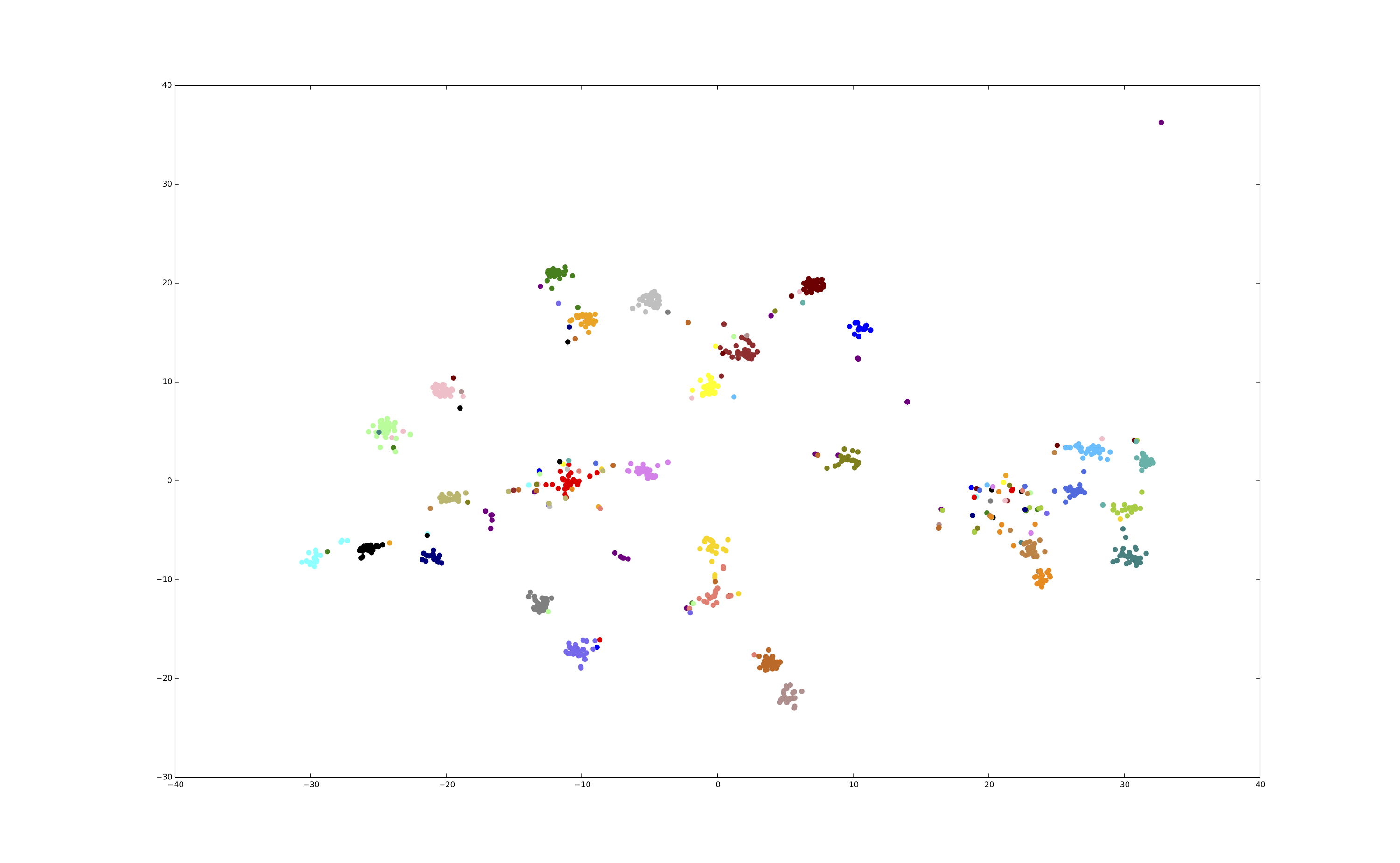}
        \caption{Proposed multi-task method}
        \label{SBHARpersPro}
    \end{subfigure}
    %\vspace*{-0.1in}
    \caption{Visualizations on the person aspect of SBHAR.}\label{SBHARpers}
\end{figure*}

In Fig.~\ref{PAMAP2act}, \ref{PAMAP2pers}, \ref{SBHARact} and \ref{SBHARpers}, different colors denote different true semantic labels in the datasets. From these figures, we can reach the following conclusions: (\rom{1}) The proposed multi-task method can effectively disentangle different semantic representations from the data. (\rom{2}) The learned representation clusters are compact and clearly separated in each representation space. Thus, downstream tasks like clustering can benefit from it\del{,} \add{and} achieve promising performance. (\rom{3}) For those datasets that contain unbalanced class numbers in different aspects of the data, the qualities of the learned representations are also unbalanced. For example, in Fig.~\ref{SBHARact} and \ref{SBHARpers}, the activity representations are more compact within the same cluster, and different clusters are clearly separated with each other. But in the person representation space, some data samples stay away from their corresponding clusters, and some clusters stay relatively close to each other. The imbalance is also reflected on the performances. As shown in Table~\ref{sbhar}, the accuracy on the activity recognition task is 10\% higher than on the person identification task. %, although both performances are better than previous works.

\begin{figure*}
    \centering
    \begin{subfigure}{0.3\linewidth}        %% or \columnwidth
        \includegraphics[width=1.13\linewidth]{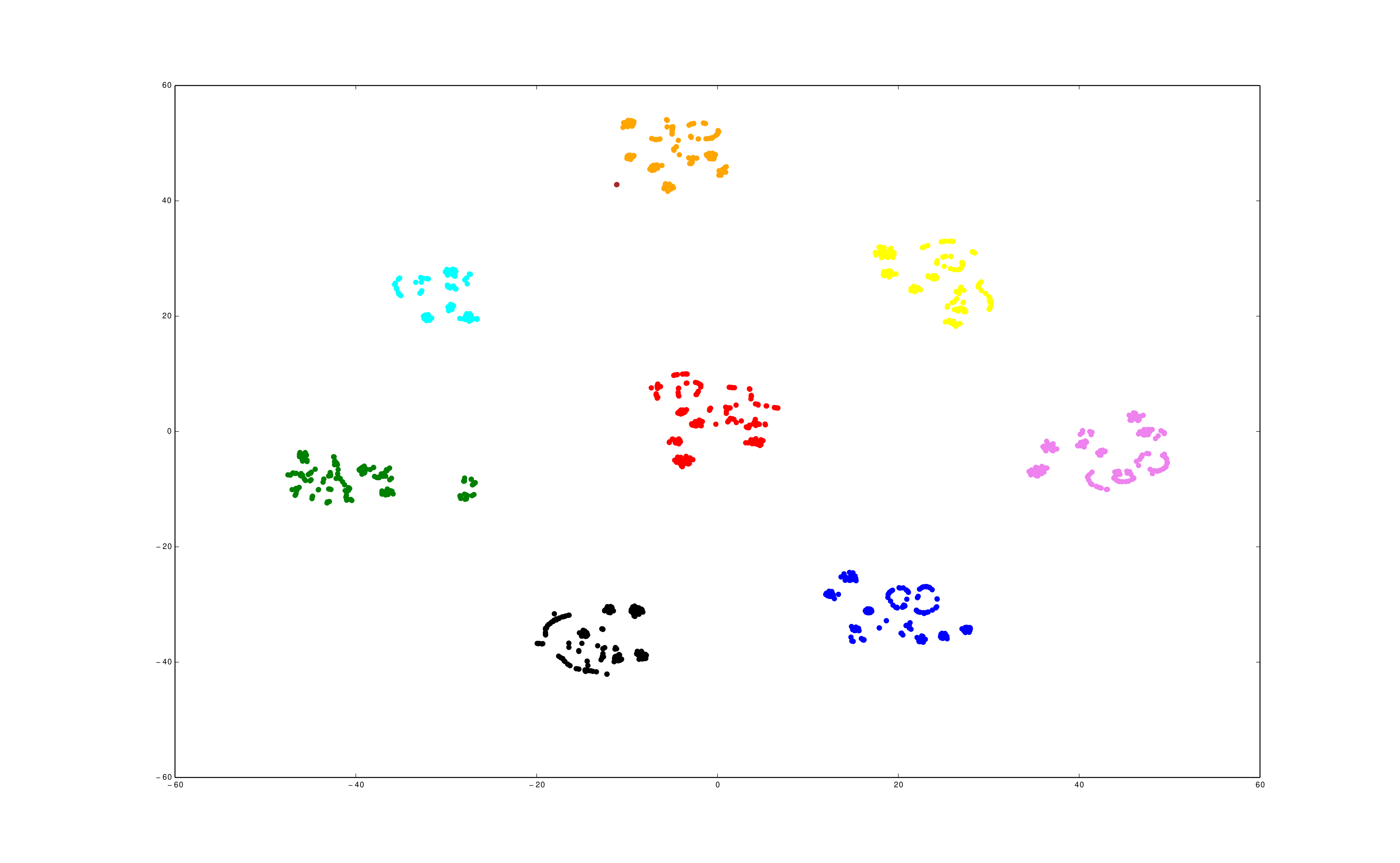}
        \caption{General rep. colored by persons}
        \label{PAMAP2gnrlpers}
    \end{subfigure}
    \begin{subfigure}{0.3\linewidth}        %% or \columnwidth
        \includegraphics[width=1.13\linewidth]{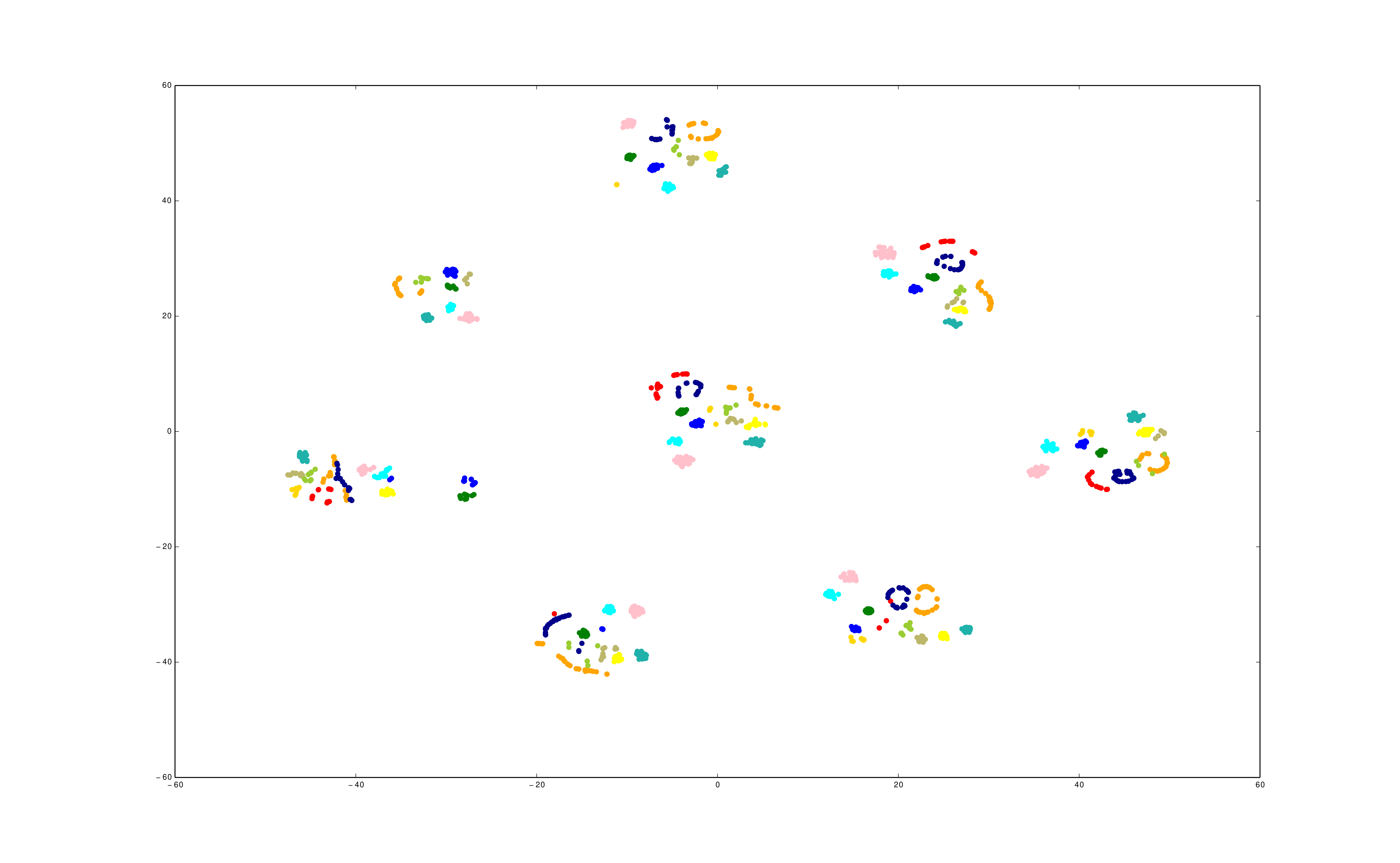}
        \caption{General rep. colored by activities}
        \label{PAMAP2gnrlact}
    \end{subfigure}
    \caption{Visualizations on the general representations of PAMAP2.} % \label{SBHARpers}
\end{figure*}

After evaluating the effectiveness of the learned representations for each task, we also visualize the output from TCN, which we consider \del{as the}\add{a more} general representation\del{. B} \add{b}ecause it is learned by the shared layers in the model without aiming at any specific task. As shown in Fig.~\ref{PAMAP2gnrlpers}, \ref{PAMAP2gnrlact}, the general representations form a two-level structure. At the higher level, \add{illustrated in} \del{Fig.~\ref{PAMAP2gnrlact}}\add{Fig.~\ref{PAMAP2gnrlpers},} the clusters are grouped in accordance with the identity of the persons. At the lower level, \add{shown in} \del{Fig.~\ref{PAMAP2gnrlpers}}\add{Fig.~\ref{PAMAP2gnrlact}}, in each person's cluster, different activities are further grouped into different smaller sub-clusters.

\hil{\subsection{Ablation Studies}
\label{ablationStudies}
As the experiment\mmadd{al} results \mmdel{shown} in Section \ref{Results} \mmadd{show}, the proposed multi-task model can learn the\mmdel{se} two related tasks (activity recognition and person identification) successfully. \mmdel{Then, a further step is to}\mmadd{To further} understand the effect of the multi-task \mmadd{training} framework,\mmdel{. A} \mmadd{a} series of ablation experiments are performed to \mmadd{separately} measure the influence of multi-task learning. In the ablation experiments, the original objective of the multi-task learning is transformed into two separate objectives \mmadd{and two separate}\mmdel{for each} single-task learning model\mmadd{s are built based on the same underlying siamese network architecture}. Thus, each single-task model is trained to learn only one task, namely activity recognition (AR) or person identification (PI). \mmdel{Moreover, each single-task model also employed the same siamese architecture as the multi-task model.}\mmadd{Comparing these to the model trained using multi-task data should provide insight into the benefit of multi-task training and thus any cross-fertilization occurring between the two tasks.} The experiment results of the ablation studies are provided in Table~\ref{ablation}.
\begin{table}[h!]
  \hil{\begin{center}
    \caption{\hil{Ablation studies on effect of multi-task learning.}}
    \scalebox{0.90}{%
    \label{ablation}
    \begin{tabular}{|c|c|c|} % <-- Alignments: 1st column left, 2nd middle and 3rd right, with vertical lines in between
      \hline
      \textbf{Methods} & \textbf{Activity} & \textbf{Person}  \\
      \hline
      \multicolumn{3}{|c|}{\textbf{PAMAP2}} \\
      \hline
%      KM & 0.6897 & 0.6360 & 0.7847 \\
      Single-Task Model AR & $ 0.9856   $ & $ - $  \\
      Single-Task Model PI & $ - $ & $0.9812  $  \\
      Proposed Multi-Task Method &$ \textbf{0.9893} $ & $\textbf{0.9881} $  \\
      \hline
      \multicolumn{3}{|c|}{\textbf{MHEALTH}} \\
      \hline
%      KM & 0.52591 & 0.37381& 0.6890 \\
      Single-Task Model AR & $0.9756   $ & $ - $    \\
      Single-Task Model PI & $ - $ & $0.9889 $    \\
      Proposed Multi-Task Method & $\textbf{0.9957}  $& $\textbf{0.9948}$  \\
      \hline
      \multicolumn{3}{|c|}{\textbf{SBHAR}} \\
      \hline
%      KM & 0.6579 & 0.5779 & 0.7188 \\
      Single-Task Model AR & $0.9436 $ & $ - $  \\
      Single-Task Model PI & $ - $ & $0.8138  $  \\
      Proposed Multi-Task Method & $\textbf{0.9885} $ & $\textbf{0.8892} $  \\
       \hline
       \multicolumn{3}{|c|}{\textbf{WISDM}} \\
       \hline
%      KM & 0.6579 & 0.5779 & 0.7188 \\
      Single-Task Model AR & $\textbf{0.9629} $ & $ - $  \\
      Single-Task Model PI & $ - $ & $0.8080  $  \\
      Proposed Multi-Task Method & $ 0.9576 $ & $\textbf{0.8112} $  \\
      \hline
    \end{tabular}}
  \end{center}}
\end{table}

As shown in Table ~\ref{ablation}, expanding the model from single-task learning to multi-task learning does not cause performance loss and degradation in most cases. On the contrary, the performance of \mmdel{ proposed} multi-task \mmadd{training}\mmdel{method} is \mmadd{most of the time better than the one of the} \mmdel{competitive to the performance of} single-task model\mmadd{s}. \mmadd{The only exception here is in the case of the WISDM dataset where a small drop in accuracy for activity recognition goes along with an increase in accuracy for person identification.} One possible reason is that the WISDM dataset is collected with one single accelerometer \mmmadd{with a relatively low sampling rate, providing significantly less data per time unit compard to the other datasets. Due to this, the time window used in this dataset is also significantly longer (10 s), leading to a much coarser grained set of labeled data points, making it potentially harder to arrange them in spaces that clearly separate them based on multiple semantic criteria due to the larger amount of overlap in the data. As a result, it might be the case here that}\mmmdel{, so} the data does not contain \mmmdel{enough}\mmmadd{sufficient} information for learning \mmmdel{a}\mmmadd{the} more complex multi-task model\mmmadd{, necessitating the observed trade-off between activity recognition and person identification}. \mmmadd{However, it is important to note here that the drop in activity recognition performance is relatively limited and smaller than the obtained gain in person identification accuracy.}
This \mmadd{result}\mmdel{also} confirms the assumption that activity recognition and person identification are closely related tasks, thus, sharing knowledge and learning a generalized representation between them can be beneficial for both tasks.}

\hil{
\subsection{\mmdel{Partially Observable} Multi-Task Learning \mmadd{with Partial Similarity Information}}
\begin{figure}
\hil{\begin{center}
\includegraphics[width=0.65\textwidth]{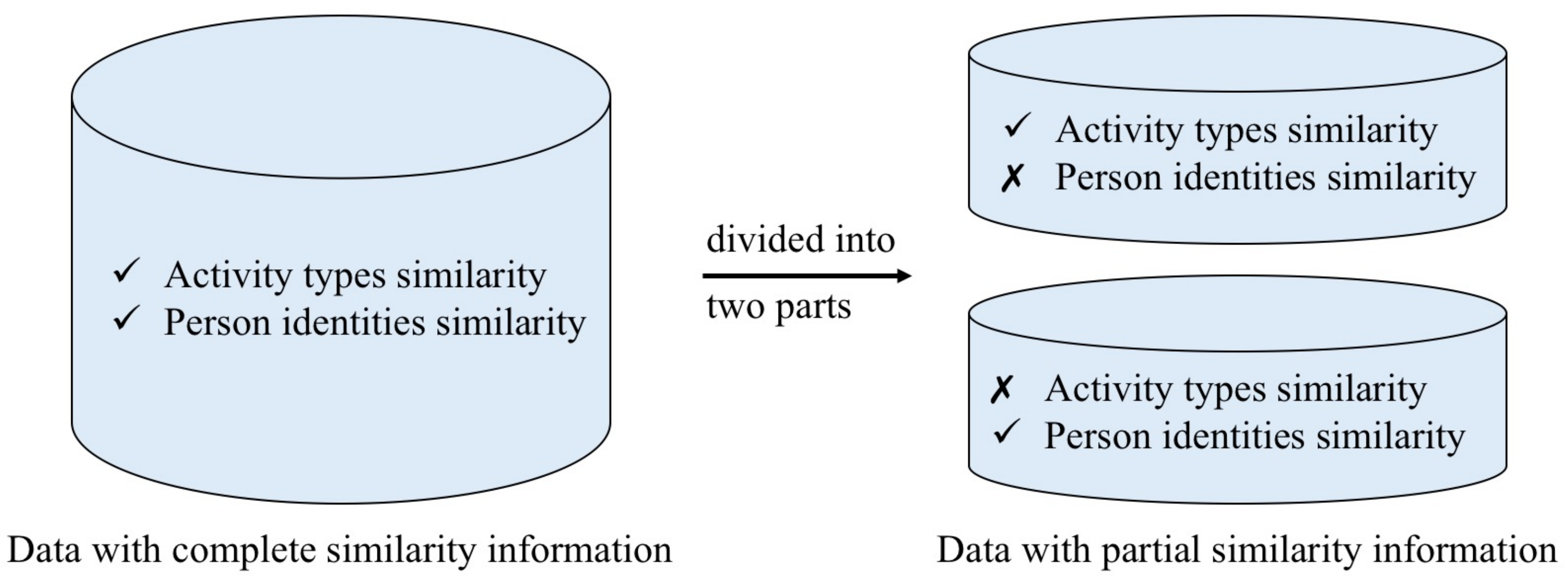} 
\caption{\hil{\mmadd{Generation of data with partial similarity information from the original data for partial information experiments}\mmdel{From complete \madd{similarity information}\mmdel{observation} to partial \mmadd{similarity information}\mmdel{observation}}.}} \label{completeIncomplete}
%\com{Is this too small ?}
\end{center}}
\end{figure}
\mmcom{The figure contains the words "obsrvable data which should be changed}
Having analyzed the effect of the multi-task framework, we now consider another\mmmadd{,} more challenging scenario. Previous experiments are all based on the assumption that the similarity information of the activity types and the person identities is fully visible to the model. However, in many real world applications\mmdel{,} it will be impossible for the model to have complete and perfect \mmdel{observation of}\mmadd{information about the similarities between all data items and for all attributes of interest}\mmdel{ the data}. Therefore, in this section, another additional constraint is imposed on the proposed multi-task method. We conducted experiments where the model is trained with \mmadd{only} partial observations of the similarity information about the data. To be more specific, as illustrated in Fig.~\ref{completeIncomplete}, we divided the dataset into two parts of equal size. One part of the data only provides the similarity information of the activity types, while the other part of the data only provides the similarity information of the person identities. \mmadd{It is important to note here that this implies that no single data item in the training set contained similarity information for both tasks and this can thus be seen as similar to a case where data was separately collected for the two tasks and no re-assessment of the similarity for the other task was performed after collection.} Then, the proposed multi-task method is trained to learn both AR and PI tasks with this \mmdel{partially observable} data. The experiment\mmadd{al} results are summarized in Table ~\ref{poMTL}.
\begin{table}[h!]
  \hil{
  \begin{center}
    \caption{\hil{\mmdel{Partially observable m}\mmadd{M}ulti-task learning \mmadd{with partial similarity information}.}}
    \scalebox{0.90}{%
    \label{poMTL}
    \begin{tabular}{|c|c|c|} % <-- Alignments: 1st column left, 2nd middle and 3rd right, with vertical lines in between
      \hline
      \textbf{Methods} & \textbf{Activity} & \textbf{Person}  \\
      \hline
      \multicolumn{3}{|c|}{\textbf{PAMAP2}} \\
      \hline
      Single-Task Model AR & $ 0.9856   $ & $ - $  \\
      Single-Task Model PI & $ - $ & $0.9812  $  \\
      Proposed Multi-Task Method with Complete \mmdel{Observation}\mmadd{Similarity Information} & $  \textbf{0.9893} $ & $ \textbf{0.9881}  $  \\
      Proposed Multi-Task Method with Partial \mmdel{Observation}\mmadd{Similarity Information} &$ 0.9849 $ & $ 0.9879 $  \\
      \hline
      \multicolumn{3}{|c|}{\textbf{MHEALTH}} \\
      \hline
      Single-Task Model AR & $0.9756   $ & $ - $    \\
      Single-Task Model PI & $ - $ & $0.9889 $    \\
      Proposed Multi-Task Method with Complete \mmdel{Observation}\mmadd{Similarity Information} & $\textbf{0.9957}  $& $\textbf{0.9948}$  \\
      Proposed Multi-Task Method with Partial \mmdel{Observation}\mmadd{Similarity Information} & $0.9744  $& $0.9643$  \\
      \hline
      \multicolumn{3}{|c|}{\textbf{SBHAR}} \\
      \hline
      Single-Task Model AR & $0.9436 $ & $ - $  \\
      Single-Task Model PI & $ - $ & $0.8138  $  \\
      Proposed Multi-Task Method with Complete \mmdel{Observation}\mmadd{Similarity Information} & $\textbf{0.9885}$ & $\textbf{0.8892}$ \\
      Proposed Multi-Task Method with Partial \mmdel{Observation}\mmadd{Similarity Information} & $ 0.9618 $ & $ 0.8247 $  \\
       \hline
       \multicolumn{3}{|c|}{\textbf{WISDM}} \\
       \hline
      Single-Task Model AR & $\textbf{0.9629} $ & $ - $  \\
      Single-Task Model PI & $ - $ & $0.8080  $  \\
      Proposed Multi-Task Method with Complete \mmdel{Observation}\mmadd{Similarity Information} & $ 0.9576 $ & $ \textbf{0.8112} $  \\
      Proposed Multi-Task Method with Partial \mmdel{Observation}\mmadd{Similarity Information}& $ 0.9509 $ & $ 0.7791 $  \\
      \hline
    \end{tabular}}
  \end{center}
  }
\end{table}

As shown in Table ~\ref{poMTL}, for both tasks, despite the data \mmdel{is} only \mmadd{containing} partial\mmdel{ly observable} \mmadd{similarity information for} \mmdel{to} the multi-task model\mmdel{,} \mmadd{and thus only providing information regarding}  \mmdel{which means that the model can learn only} one task from each sample, the performance of the \mmdel{partially observable} multi-task method is \mmdel{almost as good as}\mmadd{relatively close to} the performance of the \mmdel{completely observable} multi-task method \mmadd{using the complete similarity information for each data item}\mmdel{ in a few cases} \mmadd{and outperforms a few single-task learners trained on the data with full information from the ablation study}. \mmdel{It}\mmadd{This} indicates that our proposed method, which is trained to learn both tasks with partial\mmdel{ly observable data} \mmadd{similarity information}\mmdel{,} is capable of \mmadd{effectively utilizing cross-task information to achieve performance that is} competitive \mmdel{results} with other models trained with complete\mmdel{ly observable data} \mmadd{similarity information for all tasks}.

It is worth \mmmdel{it} to mention that one of the reasons for the model performance loss on SBHAR and WISDM is that these two datasets contain unbalanced class numbers, SBHAR contains 30 persons performing 7 activities and WISDM contains 36 persons performing 6 activities. This \mmmdel{makes}\mmmadd{leads to the situation where} each person class has significantly fewer data samples than \mmmdel{the}\mmmadd{an} activity class. Moreover, when the datasets are divided into two parts of equal size, each part only contains partial information of the data, and this further reduce\mmmdel{d}\mmmadd{s} the size of the data samples that can be used for the PI task learning. Therefore, without enough training data, the performance of the model will be degraded.
}

\subsection{Attribute Representation Learning}
\begin{figure}
\begin{center}
\includegraphics[width=0.55\textwidth]{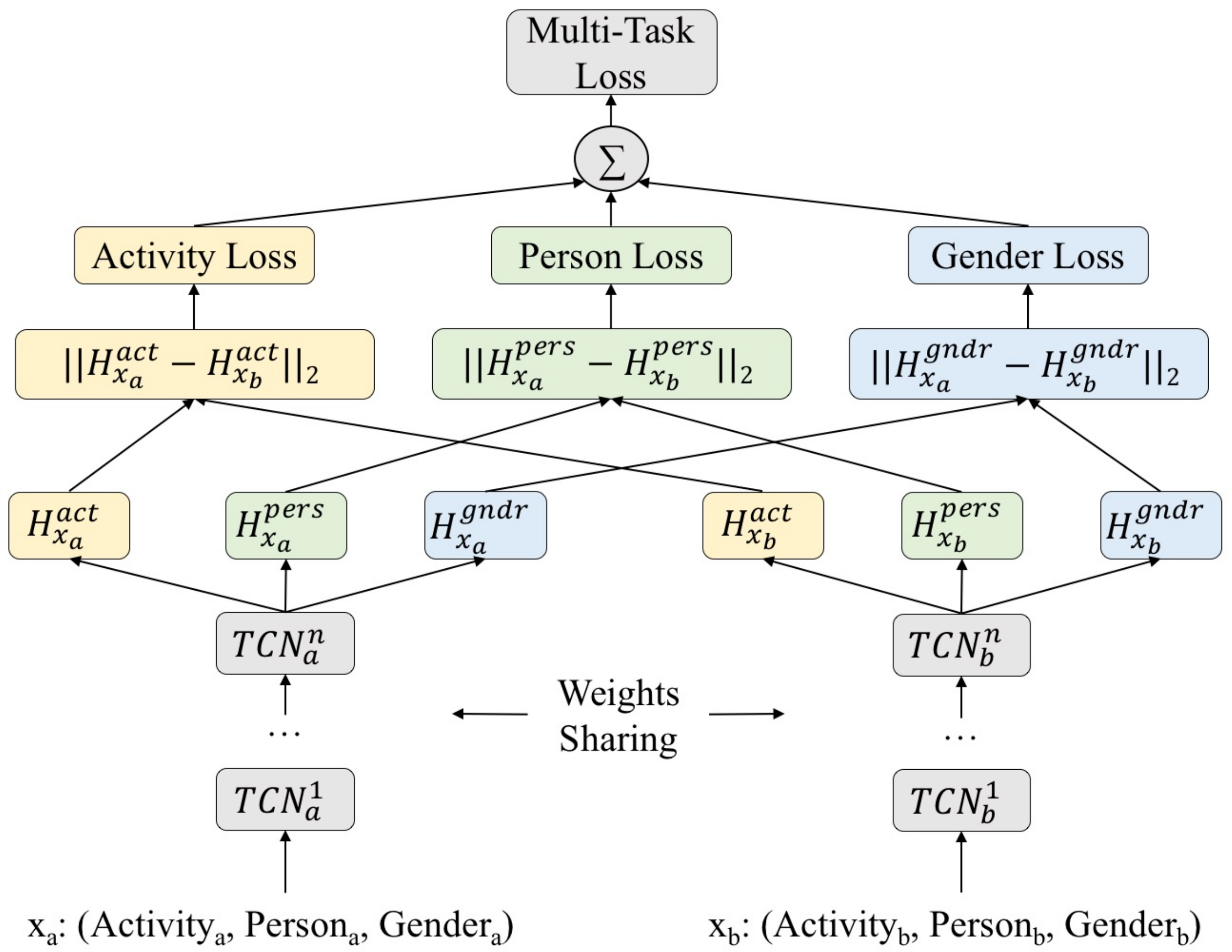}
\caption{A tri-output siamese network.} \label{triam}
\end{center}
\end{figure}

In order to evaluate the scalability and robustness of the method, we also expanded the proposed multi-task model to include additional attribute representation learning. We selected PAMAP2 to evaluate the expanded model, because this dataset provides extra attributes of the data that the model can learn. Here, we choose the gender of the person attribute as another representation learning task. The expanded model is illustrated in Fig.~\ref{triam} and shows the simplicity of expanding the proposed network to include additional representation and similarity learning tasks. 

\begin{table}
\centering
\caption{Results of proposed multi-task method on PAMAP2 with attribute representation learning in terms of $F_m$}\label{pamap23}
\scalebox{0.90}{%
\begin{tabular}{|c|c|c|c|}
\hline
\textbf{Methods} &  \textbf{Activity} & \textbf{Person} & \textbf{Gender}\\
\hline
\hil{Single-Task Model AR} & $ 0.9856   $ & $ - $  & - \\
\hil{Single-Task Model PI} & $ - $ & $0.9812  $  & - \\
\hil{Single-Task Model Gender} & $  - $ & $ - $  &  \textbf{0.8678} \\
Proposed Multi-Task Method with Dual-Output &  \textbf{0.9893} & \textbf{0.9881} & -\\
Proposed Multi-Task Method with Tri-Output &  0.9798 & 0.9827 & 0.8637 \\
\hline
\end{tabular}}
\end{table}

The experiment results are summarized in Table~\ref{pamap23}. As we can see from the table, the expanded model works well on all three tasks. Three types of different semantic representations are learned successfully. However, the unbalanced performance is also presented in this model, i.e. the performance on gender attribute learning is around 11\% lower than the performances on the other two tasks.

\section{Conclusions and Future Work}
In this paper, we propose a weakly supervised multi-output representation learning approach that is built around a siamese architecture consisting of temporal convolutions to capture multiple semantic similarities in the data simultaneously. The clustering results on the learned representations achieved promising performance and even outperformed supervised models in most situations. The visualization analysis demonstrates that the proposed approach succeeds in disentangling the semantic representations, while preserving the similarity metrics in all the representation spaces. \hil{Moreover, \mmdel{A}\mmadd{a} series of ablation studies analyzed the effect of the multi-task framework \mmadd{and showed that the use of multi-task training in this architecture most of the time improves performance over single task training, thus illustrating the frameworks ability to efficiently share information between the tasks}. \mmdel{Further a}\mmadd{A}dditional experiments showed that the proposed method can \mmadd{also} be applied to \mmdel{partially observable} data \mmadd{with only partial similarity information}, \mmadd{can be} expanded to learn new\mmadd{, additional} task easily, and achieve competitive results \mmadd{in both cases}. The comprehensive experiments give useful insights for the proposed multi-task method. Future work will be focused on designing a method that learns to adapt to the unbalancing aspects in the data automatically.}  % , and explore the capability of the model learning multi-tasks with partially observable data

\bibliographystyle{ACM-Reference-Format}
\bibliography{sample-acmlarge}

%\appendix

%Lorem ipsum dolor sit amet, consectetur adipiscing elit. Morbi
%malesuada, quam in pulvinar varius, metus nunc fermentum urna, id
%sollicitudin purus odio sit amet enim. Aliquam ullamcorper eu ipsum
%vel mollis. Curabitur quis dictum nisl. Phasellus vel semper risus, et
%lacinia dolor. Integer ultricies commodo sem nec semper.
%Etiam commodo feugiat nisl pulvinar pellentesque. Etiam auctor sodales
%ligula, non varius nibh pulvinar semper. Suspendisse nec lectus non
%ipsum convallis congue hendrerit vitae sapien. Donec at laoreet
%eros. Vivamus non purus placerat, scelerisque diam eu, cursus
%ante. Etiam aliquam tortor auctor efficitur mattis.
%Nam id fermentum dui. Suspendisse sagittis tortor a nulla mollis, in
%pulvinar ex pretium. Sed interdum orci quis metus euismod, et sagittis
%enim maximus. Vestibulum gravida massa ut felis suscipit
%congue. Quisque mattis elit a risus ultrices commodo venenatis eget
%dui. Etiam sagittis eleifend elementum.
%
%Nam interdum magna at lectus dignissim, ac dignissim lorem
%rhoncus. Maecenas eu arcu ac neque placerat aliquam. Nunc pulvinar
%massa et mattis lacinia.

\end{document}